\documentclass{article}
\usepackage{iclr2027/iclr2027_conference, times}
\usepackage{amsmath,amssymb}
\usepackage{booktabs}
\usepackage{longtable}
\usepackage{enumitem}
\usepackage{graphicx}
\usepackage{url}
\usepackage{amsthm}

\theoremstyle{definition}

\theoremstyle{plain}
\usepackage{graphicx}
\usepackage{booktabs}
\usepackage{xcolor}
\usepackage{xspace}
\usepackage{natbib}
\usepackage[colorlinks=true,allcolors=blue]{hyperref}

\newcommand{\trubric}{\texttt{trubric}}

\title{Task-efficient agent evaluation using cached traces}
\author{Ben Johnson \\
Jataware
\And
Hayden Helm\thanks{Corresponding author: \texttt{hayden@helivan.io}} \\
Helivan
}

\iclrfinalcopy

\begin{document}
\maketitle

\begin{abstract}
Efficient benchmarking estimates a system's full benchmark performance from a small number of task executions. Standard approaches rely solely on the binary correctness of executions from a collection of already-evaluated reference systems.
An agentic system, however, produces a rich trace for each execution that can reveal substantially more about its capabilities.
Leveraging this information requires useful trace representations.
We propose representing a trace by extracting a structured summary of it with an LLM and embedding the summary with a standard text encoder.
These representations preserve correctness- and task-related signal while minimizing the presence of system identity-, model family-, and harness-related signals that dominate raw trace embeddings.
Given this representation for each task execution from all reference systems, we construct low-dimensional representations of the systems and predict a new system's score from its position among the reference systems. On SWE-bench Verified and Terminal-Bench 2.0, augmenting an item-response model with this geometry-based prediction reduces mean absolute error from 0.13 to 0.08 (Wilcoxon $ p < 10^{-4} $) and from 0.14 to 0.12 (Wilcoxon $ p < 10^{-3} $), respectively, when provided a single random executed task.
Further, given access to the full set of reference systems, our approach dominates baselines for any given number of executed tasks on both benchmarks.
\end{abstract}

\section{Introduction}
\label{sec:intro}

Generative model use is increasingly agentic: systems plan, invoke tools, observe intermediate results, and iterate over long horizons to complete complex tasks \citep{yang2024sweagent, wang2025openhands}.
As agentic capabilities increase, so does the range of settings in which a given system might be deployed -- and, with it, the number of benchmarks on which the system must be evaluated before it can be trusted with consequential work \citep{kapoor2025hal}.
Agentic evaluation, however, is expensive.
Each task execution requires a properly sandboxed environment and routinely consumes hundreds of thousands of tokens \citep{jimenez2024swebench, tbench2026}.
Full evaluation on a benchmark can regularly exceed \$2{,}000 \citep{kapoor2025hal}.
Given this cost, efficient evaluation is necessary to improve the evaluation coverage of an agentic system.

Task-efficient evaluation methods address this problem by finding structure in item-level information available from previously evaluated reference systems to mitigate the cost of evaluating a new system.
Some task-efficient evaluation methods require only the item-level scores from reference systems \citep{maiapolo2024tinybenchmarks,vivek2024anchor, kipnis2025metabench, bowyer2026efficient}.
Others leverage information in cached artifacts, such as natural language responses \citep{acharyya2024consistent,helm2025statistical} or residual streams \citep{duderstadt2023data}, to measure the similarity of systems on a set of tasks.
Agentic workflows often produce artifacts when executing a task -- file diffs, tool calls, intermediate plans, and error messages -- recorded together in an execution trace that can be used to this effect.
Existing methods for using cached artifacts rely on off-the-shelf embedding functions that typically capture system or model idiosyncrasies \citep{sun2025idiosyncrasies} better than trace functionality.
Along these lines, we make two primary contributions:
\begin{enumerate}
[wide, labelwidth=1pt, labelindent=12pt]
\setlength{\itemsep}{1pt}\setlength{\parskip}{0pt}
    \item \textbf{\trubric{} trace representations.} We introduce \trubric{} (\emph{trace rubric}): a trace representation constructed by extracting a task-specific structured summary of a trace with an LLM and embedding the summary with a standard text encoder.
    We show that \trubric{} embeddings preserve information relevant to the task and behavior of the system while minimizing system-, model family-, and harness-specific signal across seven modern text encoders.
    \item \textbf{Task-efficient evaluation using cached traces.} Given the \trubric{} embeddings from a set of tasks for a target system and a collection of reference systems, we evaluate their utility for predicting the target system's full benchmark score on SWE-bench Verified and Terminal-Bench 2.0 under three different agentic system evaluation protocols.
    We find that \trubric{}-based estimators, combined with an item-response model, reduce single task mean absolute error from 0.13 to 0.08 ($p < 10^{-4}$) and from 0.14 to 0.12 ($p < 10^{-3}$) on the two benchmarks, respectively, and are nearly always better than baselines at every task budget (Table \ref{tab:hero}).
\end{enumerate}

\begin{table}[t]
\centering
\caption{Mean Absolute Error (MAE) of benchmark score estimators on SWE-bench Verified and Terminal-Bench 2.0 under leave-one-family-out evaluation and different task selection protocols. \textbf{Bolded} values are the best method for a given (benchmark, task selection, number of tasks). \underline{Underlined} values are statistically significant ($\alpha = 0.05$) comparisons to IRT per the two-sided Wilcoxon signed-rank test. The \trubric{} and IRT convex combination -- which takes advantage of both trace geometry and per-trace score -- is the best or near-best in all settings.}
\label{tab:hero}
\resizebox{\textwidth}{!}{%
\begin{tabular}{lcccccccccccc}
\toprule
 & \multicolumn{6}{c}{\textbf{SWE-bench Verified} ($n=107$)} & \multicolumn{6}{c}{\textbf{Terminal-Bench 2.0} ($n=65$)} \\
\cmidrule(lr){2-7} \cmidrule(lr){8-13}
 & \multicolumn{3}{c}{random} & \multicolumn{3}{c}{adaptive} & \multicolumn{3}{c}{random} & \multicolumn{3}{c}{adaptive} \\
\cmidrule(lr){2-4} \cmidrule(lr){5-7} \cmidrule(lr){8-10} \cmidrule(lr){11-13}
\textit{Num.\ tasks $m$} & {1} & {5} & {20} & {1} & {5} & {20} & {1} & {5} & {20} & {1} & {5} & {20} \\
\midrule
population mean & \multicolumn{6}{c}{0.139} & \multicolumn{6}{c}{0.155} \\
sample score & 0.442 & 0.175 & 0.082 & 0.380 & 0.151 & 0.072 & 0.396 & 0.164 & 0.073 & 0.360 & 0.124 & 0.132 \\
IRT & 0.127 & 0.088 & 0.048 & 0.080 & 0.062 & \bfseries 0.034 & 0.137 & 0.097 & 0.052 & 0.112 & 0.067 & 0.039 \\
raw trace & {\underline{0.099}} & 0.093 & 0.089 & 0.096 & 0.083 & 0.081 & 0.135 & 0.126 & 0.120 & 0.128 & 0.125 & 0.113 \\
generic & {\underline{\bfseries 0.079}} & {\underline{0.071}} & 0.068 & 0.074 & 0.070 & 0.069 & 0.139 & 0.127 & 0.115 & 0.144 & 0.130 & 0.117 \\
trubric & {\underline{0.081}} & {\underline{0.069}} & 0.060 & 0.068 & 0.056 & 0.060 & {\underline{0.127}} & 0.107 & 0.093 & 0.110 & 0.092 & 0.084 \\
IRT $+$ trubric & {\underline{0.081}} & {\underline{\bfseries 0.063}} & {\underline{\bfseries 0.043}} & {\underline{\bfseries 0.062}} & {\underline{\bfseries 0.048}} & \bfseries 0.034 & {\underline{\bfseries 0.122}} & {\underline{\bfseries 0.090}} & {\underline{\bfseries 0.050}} & {\underline{\bfseries 0.099}} & {\underline{\bfseries 0.059}} & \bfseries 0.037 \\
\bottomrule
\end{tabular}}
\end{table}

\subsection{Problem statement}
\label{sec:problem}

For our purposes, an agentic system $f \in \mathcal{F}$ is a random mapping from a task space $\mathcal{T}$ to a trace space $\mathcal{X}$.
A benchmark is a fixed collection of tasks $T^{*} = \{t_{1}, \hdots, t_{M}\} \subset \mathcal{T}$.
Each trace is evaluated by the benchmark's trace-level scoring function $s: \mathcal{X} \to \{0, 1\}$ and the system has a benchmark score $ y(f) = \frac{1}{M} \sum_{j=1}^{M} \mathbb{E}\left[ s(f(t_{j})) \right] $,
where the expectation is with respect to the randomness of $f$.
We write $y$ for $y(f)$ and $ y_{i} $ for $ y(f_{i}) $ when the context is clear.

When evaluating a new system, we typically have access to $ \mathcal{D} = \left\{ \{ (x_{ij}, s(x_{ij})) \}_{j=1}^{M} \right\}_{i=1}^{n} $, the traces and trace-level scores of $n$ previously evaluated reference systems $f_{1}, \hdots, f_{n}$,
where $x_{ij} \sim f_{i}(t_{j})$.
For the target system $f_{0}$, we observe traces (and, optionally, trace-level scores) on only a subset $T \subset T^{*}$ with $|T| = m \ll M$ tasks and our goal is to estimate the system's full benchmark score $ y_{0} $ using only the target's $m$ execution traces (and, optionally, their scores) and $\mathcal{D}$.

\section{Related work}
\label{sec:related}

\paragraph{Representing agentic traces.}
The standard tools for embedding model artifacts are text encoders trained on sentence-level semantic similarity \citep{reimers2019sbert, su2023instructor, nussbaum2025trainingsparsemixtureexperts}.
Agentic traces, however, implicitly or explicitly encode system metadata that may not be relevant for downstream tasks \citep{sun2025idiosyncrasies, oderinwale2026trajectories} and are therefore not necessarily encoded well by semantic embedding models.
Instead, traces have primarily been studied through taxonomies of failure modes and actions \citep{cemri2025mast, shu2026traceprobe} and LLM judges \citep{zhuge2025agentjudge, mehtiyev2026beyond}.
Parallel work uses LLMs to adapt outputs to be more naturally compared via their semantics \citep{ravfogel2024description, peng2024inbedder, benara2024qaemb}.
\trubric{} applies this to agentic traces by embedding a task-specific structured summary of each trace.

\paragraph{Efficient benchmarking.}
Task-efficient benchmarking estimates a system's full benchmark score from a small number of executions.
The dominant paradigm for efficient benchmarking centers around item response theory (IRT) \citep{birnbaum1968some}, where a target system's score is estimated with $ m \le M $ item-level scores by assigning a latent ability based on the relationship between a task's difficulty and discriminability \citep{maiapolo2024tinybenchmarks, kipnis2025metabench}.
Extensions of IRT select informative or adaptive task subsets \citep{vivek2024anchor, truong2025amortized}.
A complementary paradigm represents the systems themselves, either from their score profiles \citep{ruan2024observational, zhuang2025embedllm} or from the artifacts they produce while executing a task \citep{duderstadt2023data, acharyya2024consistent, helm2025statistical}.
For example, the Product Kernel Perspective Space (PKPS) \citep{helm2026benchmarkpredictionsparseunpaired} represents language models as Euclidean objects obtained via multidimensional scaling of the distances between their embedded responses.
\citet{helm2026query} regress benchmark scores on similar representations and prove that the estimate is task-efficient relative to the sample score.
\trubric{} enables extending geometry-based methods to agentic systems by providing an effective trace encoding.
We show that these two approaches -- score-based IRT and trace-based PKPS -- can be combined to outperform either constituent for any given number of executed tasks when the full set of reference systems is available.

\section{Methods}
\label{sec:method}
We first present \trubric{}.
We then describe constructing representations of the systems $ f_{0}, f_{1}, \hdots, f_{n} $ based on \trubric{} embeddings that can be used to train a simple regressor in $ \mathbb{R}^{d} $ for task-efficient evaluation.
We evaluate the proposed methods in Section \ref{sec:experiments}.

\subsection{Task-specific rubrics for representing traces}
\label{sec:method-trubric}
\begin{figure}[!b]
\centering
\includegraphics[width=\textwidth]{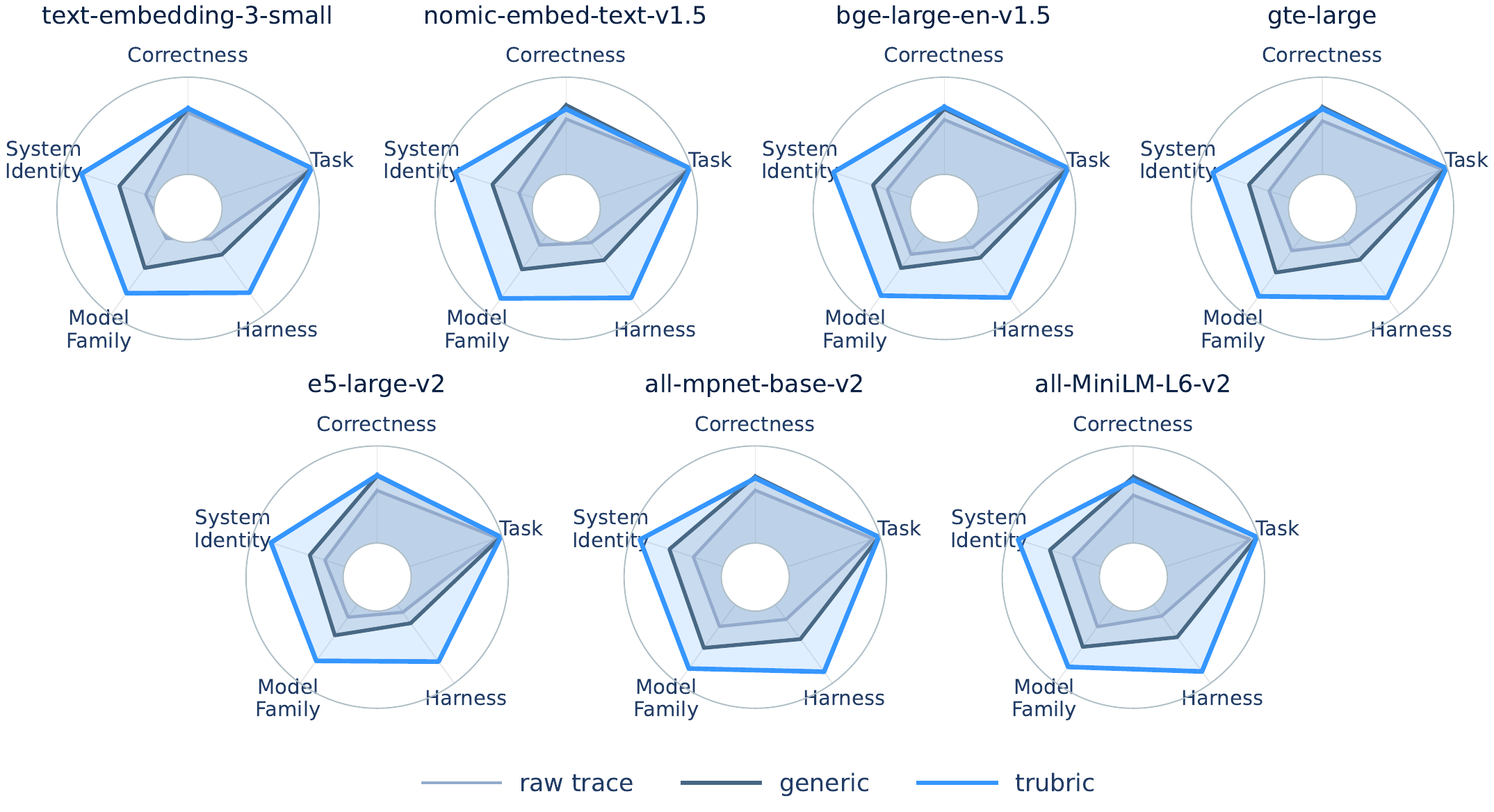}
\caption{Three different methods for representing traces (embedding the raw trace, embedding a structured summary based on a generic rubric, and embedding a structured summary based on a task-specific rubric) evaluated on preserving / removing attributes (i)--(v).
The representation based on \trubric{} is better at preserving relevant signal ((i) Task and (ii) Correctness) while mitigating information about system idiosyncrasies ((iii) System Identity, (iv) Model Family, and (v) Harness).}
\label{fig:radar}
\end{figure}

A trace representation for benchmark prediction should place two traces close together when they reflect similar execution behavior on a task.
Concretely, we consider five attributes of a trace that a representation might encode: (i) task (i.e., which task the trace corresponds to), (ii) correctness (i.e., whether the trace produced a correct output), (iii) system identity (i.e., which system produced the trace), (iv) model family (i.e., the family of the underlying generative model), and (v) harness (i.e., the harness used to produce the trace).
For efficient benchmarking with traces, an ideal representation encodes (i) and (ii) but not (iii)--(v).
\trubric{} constructs such a representation in two stages.

First, for each task $t_{j}$, an LLM is prompted with the task description to instantiate a task-specific rubric.
The rubric is a set of $K$ fields $R_{j} = (r_{j}^{(1)}, \hdots, r_{j}^{(K)})$ where each field describes an aspect of execution behavior (e.g., how the relevant code was found).
Rubrics are shared across all traces for a given task and fields are shared across tasks.
An example rubric is provided in Appendix \ref{app:example}.

Second, for each trace $x_{ij}$ we use an LLM to produce a short summary $z_{ij}^{(k)} $ for each field.
The summary of each field is then embedded separately with a semantic embedding model $g$ mapping text to $\mathbb{R}^{p}$ and concatenated to yield the trace representation $\phi(x_{ij}) = ( g(z_{ij}^{(1)}), \hdots, g(z_{ij}^{(K)})) \in \mathbb{R}^{K \cdot p} $.
Two examples of an extracted summary for a given task are provided in Appendix \ref{app:example}.

Figure \ref{fig:radar} compares \trubric{} to two baselines along axes (i)--(v).
Values closer to the edge correspond to a more ideal representation for efficient benchmarking for a given attribute.
Experiment details are provided in Section \ref{sec:experiments}.

\subsection{Inference in the Product Kernel Perspective Space}
\label{sec:method-pkps}

Given trace representations, we represent agentic systems via the Product Kernel Perspective Space (PKPS).
Before constructing the geometry of systems, trace representations are centered by subtracting the average across systems for each task and field to emphasize differences in behavior.
Let $\tau$ denote a text embedding of the task description and $\kappa_{\sigma}(t, t') = \exp\left(-\| \tau(t) - \tau(t') \|^{2} / 2\sigma^{2}\right)$.
Let $ \langle \phi, \phi'\rangle = \phi^{T} \phi'$ for $ \phi, \phi' \in \mathbb{R}^{K \cdot p} $.
For systems $f_{a}$ and $f_{b}$ evaluated on task sets $T_{a}$ and $T_{b}$, define the weighted similarity
\begin{equation}
\label{method:pkps}
\langle f_a, f_b \rangle = \frac{\sum_{t \in T_{a}} \sum_{t' \in T_{b}} \kappa_{\sigma}(t, t') \, \langle \phi(x_{at}), \phi(x_{bt'}) \rangle}{\sum_{t \in T_{a}} \sum_{t' \in T_{b}} \kappa_{\sigma}(t, t')},
\end{equation}
the corresponding squared distance $\delta^{2}(f_a, f_b) = \langle f_a, f_a \rangle + \langle f_b, f_b \rangle - 2 \langle f_a, f_b \rangle$, and $ \Delta_{ab} = \delta^{2}(f_a,f_b) $. With $ \Delta $ defined on $ f_{0}, f_{1}, \hdots, f_{n} $, applying classical multidimensional scaling \citep{torgerson1952mds} to $ \Delta $ yields $d$-dimensional Euclidean representations $(\psi_{0}, \psi_{1}, \hdots, \psi_{n}) \in \mathbb{R}^{(n+1) \times d}$ of the systems with respect to the collective task sets.
Importantly, Eq. \eqref{method:pkps} is defined when $ |T_{a} | \neq |T_{b}| $ if both $ |T_{a}| > 0 $ and $ |T_{b}| > 0 $, so the target system is naturally represented in the same space as the reference systems.
It is typical that $ |T_{0} | = m \ll M $ and $ |T_{i} | = M $ for all $ i \in \{1, \hdots, n\} $.

\begin{figure}[t]
\centering
\includegraphics[width=\textwidth]{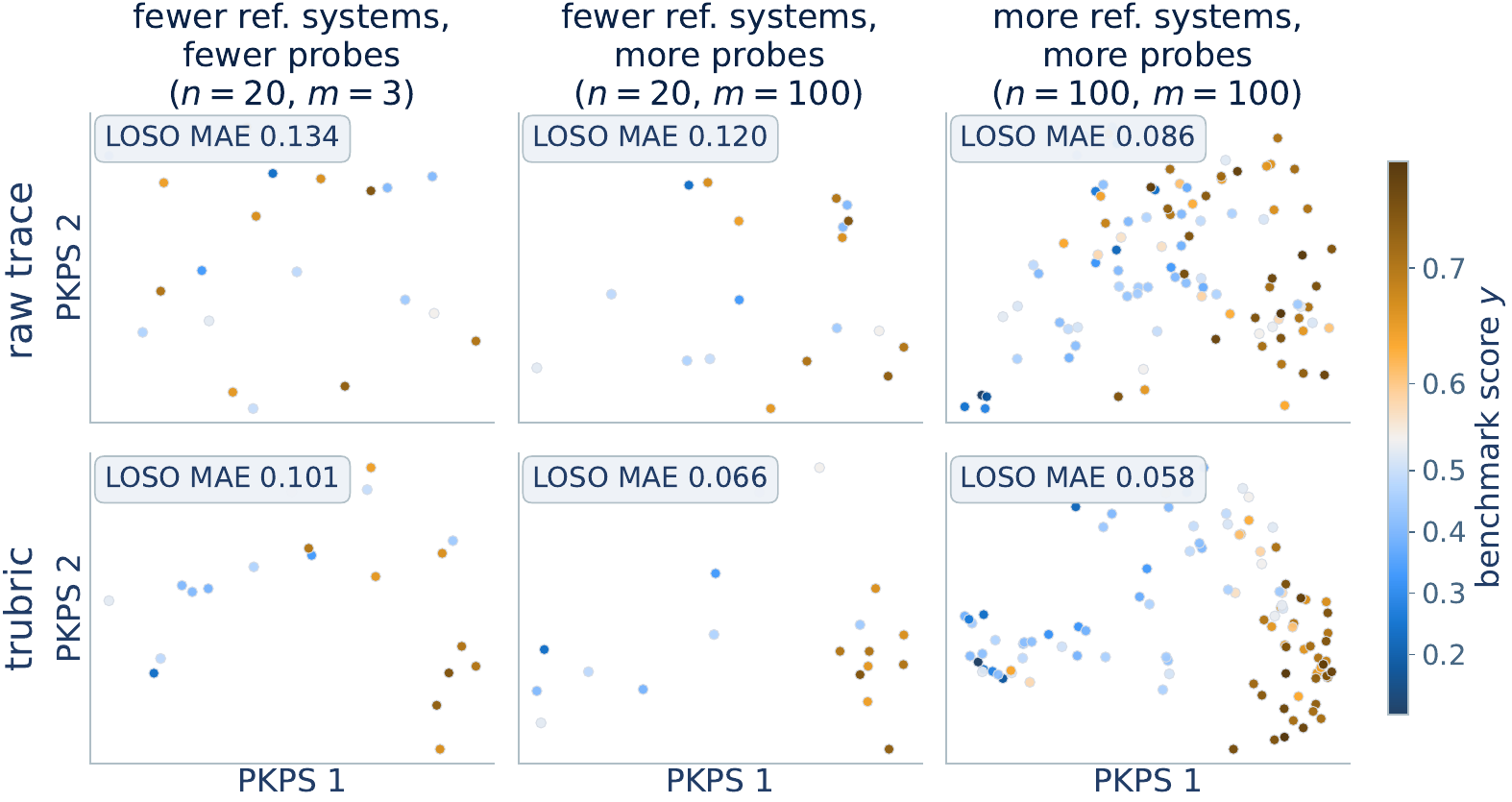}
\caption{The 2-d Product Kernel Perspective Space (PKPS) induced with raw trace embeddings (top) and the \trubric{} embeddings (bottom) under different numbers of reference systems and task executions.
Each dot is a system colored by its full benchmark score on SWE-bench Verified.
For a given setting, the manifold of systems induced by \trubric{} embeddings is more predictive of the full benchmark score, as measured by MAE under the leave-one-system-out protocol.}
\label{fig:pkps}
\end{figure}

Given $ \mathcal{D} $ and $ \psi_{1}, \hdots, \psi_{n} $, we train a simple regressor on $ (\psi_{1}, y_{1}), \hdots, (\psi_{n}, y_{n})$ and apply it to $ \psi_{0} $ to obtain $ \hat{y}_{\mathrm{geom}} $, an estimate of $ f_{0} $'s full benchmark score $ y_{0} $.
Figure \ref{fig:pkps} shows the $ d=2$ perspectives when using the raw-trace embeddings (top) and the \trubric{} embeddings (bottom) for different numbers of reference systems and numbers of executed tasks per system.
Each dot is a system colored by its full benchmark score on SWE-bench Verified.
The \trubric{}-based PKPS contains more information, per a lower mean absolute error on leave-one-system-out evaluation (LOSO; see Section~\ref{sec:experiments}).
Finally, we note that $ \hat{y}_{\mathrm{geom}} $ does not use any item-level score information.

\section{Experiments}
\label{sec:experiments}

We next introduce the datasets we use to evaluate the trace embeddings and the trace-based geometries for efficient evaluation, describe the trace embedding analysis and three different agent evaluation protocols, and describe baselines.
We then evaluate \trubric{}-based estimators against baselines across benchmarks, evaluation protocols, task budgets, and design choices.

\paragraph{Benchmark data.}
We consider two agentic benchmarks:
\begin{enumerate}
[wide, labelwidth=1pt, labelindent=12pt]
\setlength{\itemsep}{1pt}\setlength{\parskip}{0pt}
    \item \textbf{SWE-bench Verified} \citep{jimenez2024swebench, openai2024verified} consists of 500 human-validated software-engineering tasks drawn from real GitHub issues.
    Its public leaderboard includes submissions from 119 unique systems.
    We consider the 107 systems with complete public trace archives and per-task outcomes.
    We downsampled to $ M = 100 $ tasks (Appendix~\ref{app:data}).
    We let $ y_{i} $ be the official resolve rate over all 500 tasks.
    The full list of systems and tasks is included in Appendix~\ref{app:data}.
    \item \textbf{Terminal-Bench 2.0} \citep{tbench2026} consists of 89 command-line tasks executed under a standardized harness specification.
    Its leaderboard contains submissions from 75 systems.
    Each system attempts each task five times.
    We analyze the 65 unique systems with at least 90\% public trace coverage.
    We use a single trial's trace and binary outcome for a given (system, task) and let $ y_{i} $ be the mean resolve rate across all tasks and all trials.
    The full list of systems and tasks is included in Appendix~\ref{app:data}.
\end{enumerate}

\paragraph{Trace embedding evaluation.}
To quantify what a trace representation encodes, we train balanced linear classifiers to predict each of the five attributes of Section~\ref{sec:method-trubric} for a held-out trace from its embedding.
Classifiers for the system-level attributes (iii)--(v) are evaluated on held-out tasks.
Classifiers for (i) and (ii) are evaluated on held-out groups of systems.
We evaluate trace representations using the cached SWE-bench Verified traces and report balanced accuracy for each attribute.
Higher accuracy is preferred for attributes (i) and (ii).
Near chance accuracy is preferred for attributes (iii)--(v).
Detailed descriptions of the classification tasks are provided in Appendix~\ref{app:probes}.

\paragraph{Trace embedding baselines.}
We compare \trubric{} embeddings to two baseline representations: \emph{raw trace}, which uses $ g $ to embed the start and end of each trace, and \emph{generic}, which follows the two-stage construction of Section~\ref{sec:method-trubric} with fixed field descriptions in place of a task-specific rubric.
Unless otherwise stated, both the generic and \trubric{} representations use $K = 6$ fields, with DeepSeek-V4-Flash \citep{deepseek2026v4} as both the rubric generator and summarizer.
For the embedding function $g$, we consider seven text encoders: \texttt{text-embedding-3-small} \citep{neelakantan2022text}, \texttt{nomic-embed-text-v1.5} \citep{nussbaum2024nomic}, \texttt{bge-large-en-v1.5} \citep{xiao2023cpack}, \texttt{gte-large} \citep{li2023gte}, \texttt{e5-large-v2} \citep{wang2022e5}, \texttt{all-mpnet-base-v2}, and \texttt{all-MiniLM-L6-v2} \citep{reimers2019sbert}.
These encoders span open- and closed-source models, multiple architectures and training procedures, and embedding dimension sizes.
The prompts and fixed field descriptions are provided in Appendix~\ref{app:prompts}.

\paragraph{Efficient benchmarking evaluation protocols.}
Modern leaderboards contain many entries that share an underlying model or harness.
To emulate the process of evaluating a new model or harness, as well as a new combination of the two, we consider three different evaluation protocols: \emph{leave-one-system-out} where only the target itself is excluded from the references,
\emph{leave-one-family-out}  where references whose underlying model family matches the target's are additionally excluded, and \emph{leave-one-harness-out} where references sharing the target's harness are excluded.
Detailed descriptions of a system's assignment to a model family or harness are provided in Appendix \ref{app:data}.

\paragraph{Efficient benchmarking baselines.}
We compare \trubric{}-based predictions against five baselines: the \emph{population mean} (the mean benchmark score of the reference systems), the \emph{sample score} (the mean of the target's $m$ observed trace-level scores), \emph{raw trace} (the PKPS pipeline applied to the raw-trace embeddings), \emph{generic} (the PKPS pipeline applied to the generic rubric embeddings), and two-parameter \emph{IRT}.
We also consider the convex combination $ \hat{y}_{\alpha} = \alpha \, \hat{y}_{\mathrm{IRT}} + (1 - \alpha) \, \hat{y}_{\mathrm{geom}}$, with $\alpha \in [0, 1] $ selected via cross-validation.
We refer to the convex combination as IRT $+$ \trubric{}.
For \trubric{}, raw trace, and generic, the regressor and its hyperparameters are selected per task draw via cross-validation on the reference systems. Implementation details for all estimators, including a technical description of IRT, are provided in Appendix~\ref{app:baselines}.

\paragraph{Task selection methods.}
For each task budget $m$, we consider two methods for selecting the target's $m$ executed tasks.
\emph{Random} selection draws the $m$ tasks uniformly. Reported results average over 50 randomly sampled task subsets.
\emph{Adaptive} selection chooses tasks greedily using the fitted item-response model. In particular at each step, the task with the highest expected Fisher information under the current ability posterior is executed and the posterior is updated (Appendix~\ref{app:baselines}).

\subsection{Results}
\label{sec:results}
\begin{figure}[t]
\centering
\includegraphics[width=\textwidth]{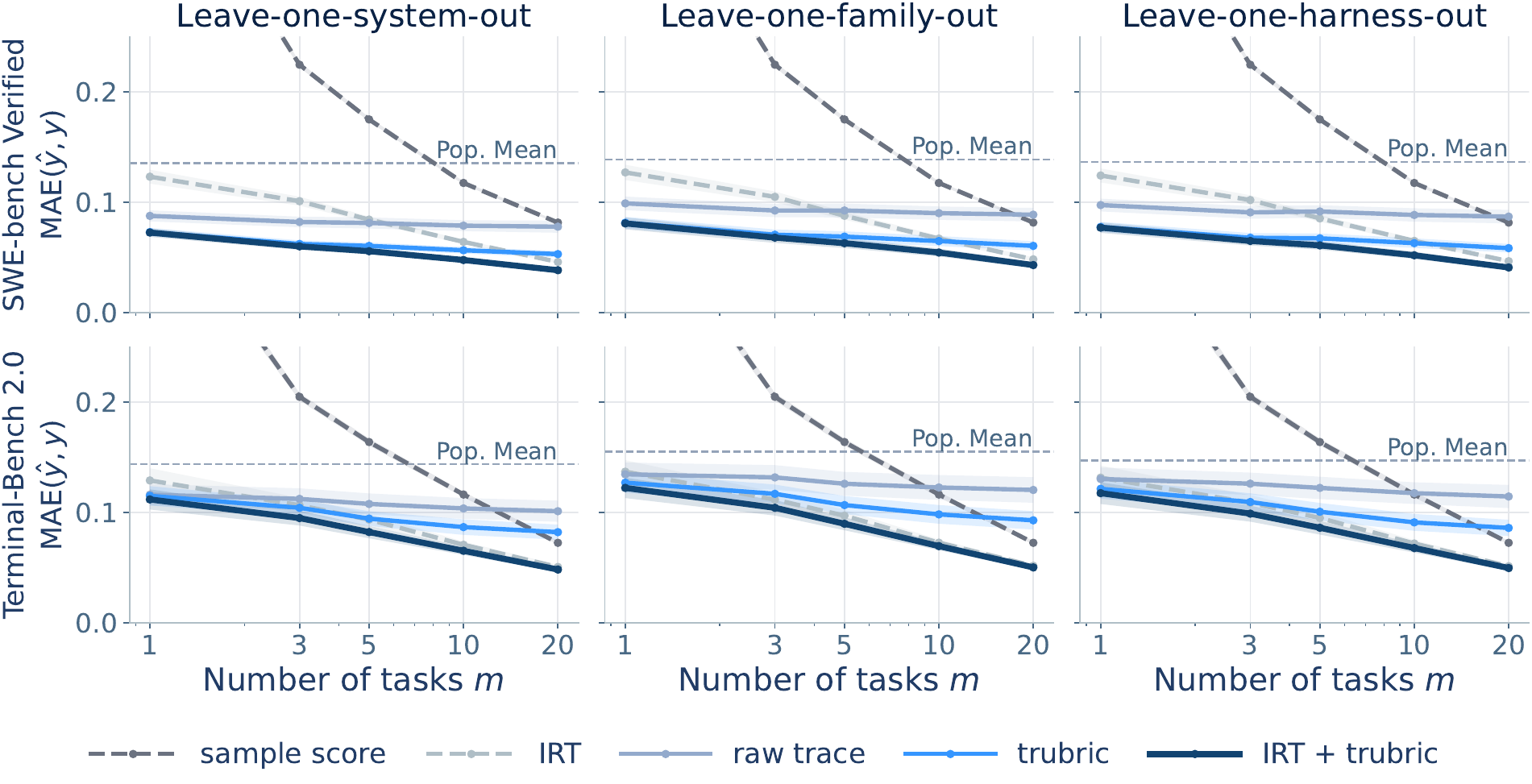}
\caption{MAE of benchmark score estimators as a function of the number of executed tasks for SWE-bench Verified (top) and Terminal-Bench 2.0 (bottom) under three different realistic evaluation protocols: leave-one-system-out (left), leave-one-family-out (center), and leave-one-harness-out (right).
Shaded areas represent $ \pm 1 $ S.E.
The \trubric{} and IRT combination dominates (i.e., is never worse and is sometimes better) across all protocols, benchmarks, and numbers of executed tasks.}
\label{fig:protocols}
\end{figure}

\paragraph{Rubric-based preprocessing produces more useful trace representations.}
Figure~\ref{fig:radar} shows that both generic and task-specific rubrics reduce recoverable information about system identity, model family, and harness while retaining task and correctness information.
As discussed above, these attributes can introduce nuisance variation and obscure similarities in how systems execute tasks.
Task-specific rubrics produce a larger reduction in identity-related signals compared to generic rubrics.
These qualitative patterns persist across all seven embedding models.

The differences in attributes (i)--(v) are also reflected in the quality of the system-level geometry for predicting benchmark score.
Figure~\ref{fig:pkps} shows a stronger relationship between system position in the embedding space and benchmark score for \trubric{} than for raw-trace embeddings.
In Figure~\ref{fig:protocols}, leave-one-system-out prediction error decreases more rapidly with additional tasks when using \trubric{} representations than when using raw-trace embeddings.
On SWE-bench Verified under leave-one-family-out evaluation, increasing the task budget from 1 to 20 reduces geometry-only MAE from 0.081 to 0.060 for \trubric{}, compared with 0.099 to 0.089 for raw traces.
Table~\ref{tab:hero} further shows that generic rubrics are competitive on SWE-bench at small budgets, though task-specific rubrics provide larger improvements on Terminal-Bench 2.0.
For the remainder of the paper ``trace geometry'' refers to a \trubric{}-based geometry.

\paragraph{Trace geometry supports task-efficient benchmark prediction.}
Table~\ref{tab:hero} and Figure~\ref{fig:protocols} show that \trubric{}-based geometry predicts full benchmark performance more accurately than the sample score for small task budgets ($ m \le 10 $) and the population mean for all budgets.
The \trubric{}-based geometry can also outperform IRT, which exploits the correctness patterns of reference systems, at small task budgets.
On SWE-bench, the geometry produces a more accurate estimate from 1 through 10 random tasks (Figure~\ref{fig:protocols}).
On Terminal-Bench, $\hat{y}_{\mathrm{geom}} $ outperforms IRT at one task, but $ \hat{y}_{\mathrm{IRT}} $ is more accurate from three tasks onward.
We emphasize that the geometric methods \emph{only} require the target's trace content, and \emph{not} its correctness labels, which means that these methods are applicable in settings where grading target tasks is infeasible (e.g., due to cost, lack of expertise, etc.), provided scored reference systems are available.
$ \hat{y}_{\mathrm{geom}} $ can be interpreted as a high-bias, low-variance estimate of the full benchmark score.

\begin{figure}[t]
\centering
\includegraphics[width=\textwidth]{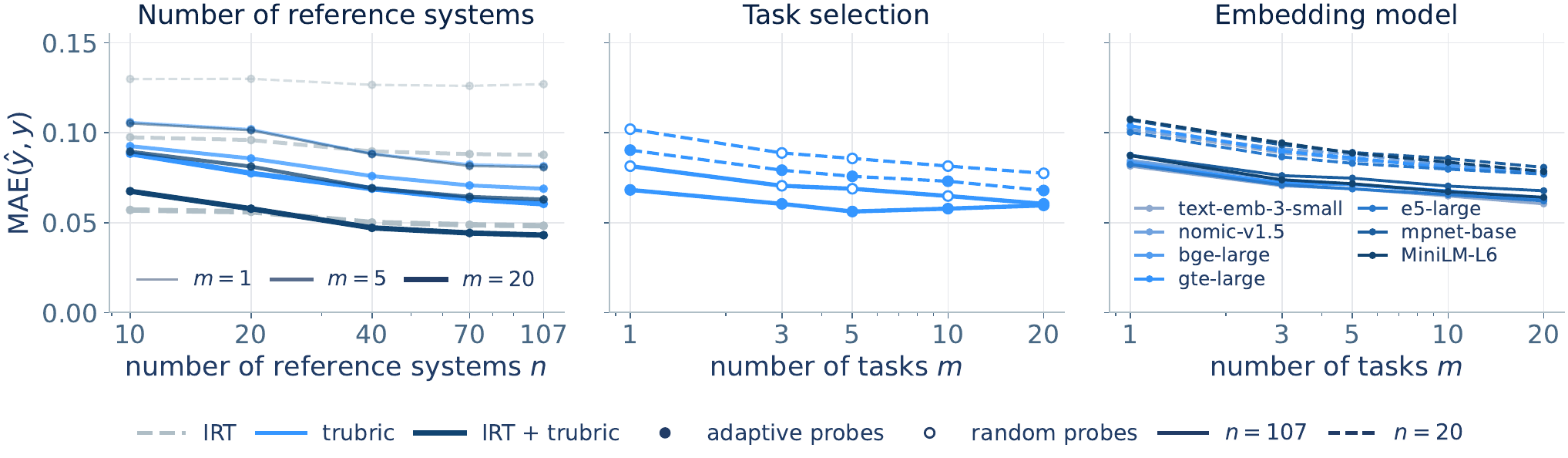}
\caption{MAE of benchmark score estimators with access to different numbers of reference systems (left), under different task selection procedures (center), and using different text encoders (right) under the leave-one-family-out evaluation protocol on SWE-bench Verified.
The embedding model analysis (right) only investigates the effect of the embedding model on $ \hat{y}_{\mathrm{geom}} $.}
\label{fig:ablations}
\end{figure}

\paragraph{Combining trace geometry with IRT is best.}
The trace-based predictions and correctness-based predictions exhibit different behavior as the task budget grows: trace-based predictions are informative with very few executions, while IRT improves substantially as the number of tasks increases.
Figure~\ref{fig:protocols} shows that combining their predictions improves prediction across all evaluated budgets and benchmarks.
At $ m =5 $ on SWE-bench, IRT $+$ \trubric{} achieves 0.063 MAE, compared with 0.069 for geometry alone (Wilcoxon's $ p < 10^{-4} $) and 0.088 for IRT (Wilcoxon's $ p < 10^{-12} $).
On Terminal-Bench -- where IRT is relatively stronger -- the combination achieves 0.090, compared with 0.107 for geometry (Wilcoxon's $ p < 10^{-3} $) and 0.097 for IRT (Wilcoxon's $ p < 10^{-2} $).
These results suggest a complementary predictive value of execution artifacts and correctness outcomes.
The broad pattern persists across the system-, family-, and harness-exclusion protocols in Figure~\ref{fig:protocols}, although the size of the improvement varies.

\paragraph{Performance improves with larger reference sets.}
The left panel of Figure~\ref{fig:ablations} shows that increasing the number of reference systems generally improves both IRT and geometry-based estimators.
At $ m = 5 $ random tasks on SWE-bench, expanding from $ n = 20 $ to $ n = 107 $ reduces geometry-only MAE from 0.086 to 0.069 and IRT $+$ \trubric{} MAE from 0.081 to 0.063.
The contribution of geometry depends on reference availability: given a small number of reference systems and a large task budget, IRT can outperform geometry-only methods.
The medium budget ($ m = 20 $), small-reference-set ($n=20$) regime is the one setting where IRT outperforms the convex combination.

\paragraph{Adaptive task selection composes with trace-based estimation.}
Adaptive task selection provides a further improvement.
With $ m = 5 $ adaptive tasks, IRT $+$ \trubric{} achieves 0.048 MAE on SWE-bench, compared with 0.063 for five random tasks (Table~\ref{tab:hero}).
Together, Table~\ref{tab:hero} and the center panel of Figure~\ref{fig:ablations} show that adaptive selection improves both IRT and geometry-based prediction and can further improve evaluation efficiency.

\paragraph{Prediction is insensitive to the choice of text encoder.} We evaluate the sensitivity of \trubric{}-based prediction on SWE-bench to the choice of text encoder in the right panel of Figure~\ref{fig:ablations}.
The effect is small (at most 0.008 MAE at any fixed setting) relative to the effect of the number of target tasks or task selection (0.081 to 0.060 from 1 to 20 random tasks; 0.063 to 0.048 for random versus adaptive selection at $ m = 5$) and the number of reference systems (0.086 to 0.069 at $ m =5$).

\begin{figure}[t]
\centering
\includegraphics[width=\textwidth]{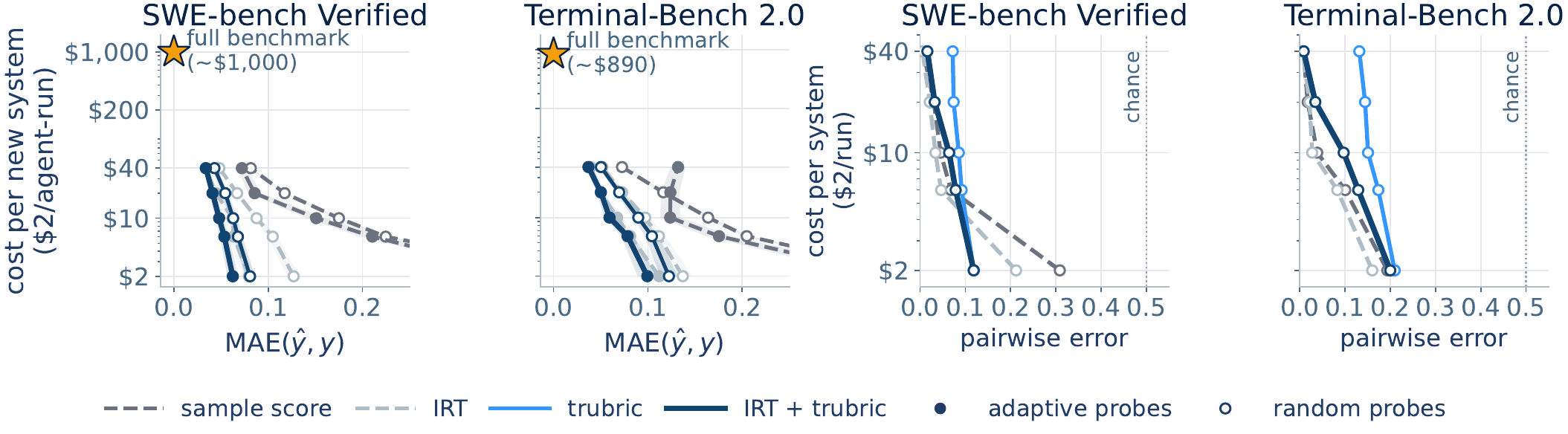}
\caption{The cost of achieving a particular MAE (left and center left) and pairwise ranking error (center right and right) when using different benchmark score estimators on SWE-bench Verified and Terminal-Bench 2.0. Stars indicate the cost of producing the true benchmark score. An ideal method has low error (to the left) and is cheap (lower). Costs are calculated at a nominal \$2 per task execution. Values are estimated using the leave-one-family-out results underlying Table~\ref{tab:hero} and, for pairwise error, leave-two-out comparisons of system pairs whose true scores differ by at least 0.05.}
\label{fig:cost}
\end{figure}

\paragraph{The trace geometry reduces the cost of reaching a given prediction error.}
Figure~\ref{fig:cost} translates the error curves into the cost of evaluating a system.
For this comparison, we assume a cost of \$2 per agent execution \citep{jimenez2024swebench, tbench2026, kapoor2025hal} and assume both the cost of using LLMs to do (t)rubric creation and summarization and the cost of embedding the traces are negligible.
On SWE-bench, IRT $+$ \trubric{} reaches an MAE below 0.10 with one random task, compared with five tasks for IRT and twenty for the sample score.
This corresponds to estimated target execution costs of \$2, \$10, and \$40, respectively.

We also evaluate methods for comparing two systems head to head.
For each pair of systems whose true benchmark scores differ by at least 0.05, we exclude both systems from the references and ask whether an estimator correctly identifies the better system from $m$ random executions per system (right two panels of Figure~\ref{fig:cost}).
On SWE-bench, IRT $+$ \trubric{} decides 88\% of pairs correctly from a single task per system, compared with 79\% for IRT and 69\% for the sample score.
By five tasks per system, all estimators exceed 90\% and the score-based estimators are competitive or better.
On Terminal-Bench, IRT is the most accurate estimator at every tested budget.
Trace information is thus most valuable for head-to-head comparisons in the lowest budget regime, where executing a single task per system (\$4 for the pair) identifies the better system in nearly nine of ten pairs.

\begin{figure}[t]
\centering
\includegraphics[width=\textwidth]{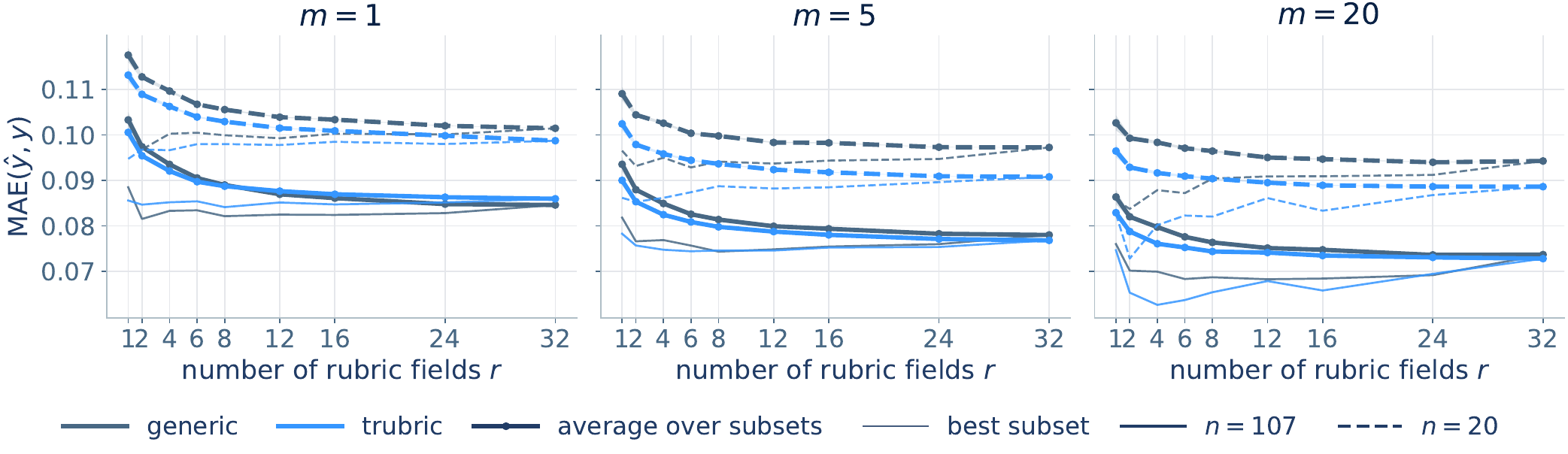}
\caption{The effect of the number of fields present in the rubric used to extract a structured summary of the trace on MAE for different numbers of executed tasks and reference systems for both task-specific and generic rubrics.
We report both the MAE for the average (bold) and best performing (thin) sets of fields.
In general, a task-specific rubric is preferred to a generic rubric when there are fewer reference systems and more executed tasks available.}
\label{fig:sensitivity}
\end{figure}

\paragraph{Larger rubrics improve average performance.}
Figure~\ref{fig:sensitivity} varies the number of fields included in generic and task-specific rubrics, averaging performance over subsets drawn from a bank of 32 fields.
For both constructions, adding fields generally improves average prediction accuracy.
At $ m = 5 $ and a large set of reference systems, increasing a task-specific rubric from one to six fields reduces average MAE from 0.090 to 0.081 (Wilcoxon's $ p < 10^{-5} $).
Using all 32 fields reduces it further to 0.077 (Wilcoxon's $ p < 10^{-3} $).
Task-specific rubrics generally outperform generic rubrics, especially with fewer reference systems and larger number of executed tasks.

Figure~\ref{fig:sensitivity} also reports the best performing rubric.
Each point selects the lowest average error among the tested subsets separately for that setting.
These \textit{post hoc} selections demonstrate the existence of favorable small rubrics, which may be preferred in settings where embeddings of the traces are stored for long periods.
Importantly, the selected subsets need not be nested as field count increases, so their performance need not improve monotonically.
We did not attempt to optimize the fields.
The full list of the 32 field names is provided in Appendix~\ref{app:fields}.

\section{Discussion}
\label{sec:discussion}

We introduced \trubric{}, a trace representation constructed by extracting a task-specific structured summary of a trace with an LLM and embedding the summary with a standard text encoder.
Across seven text encoders, \trubric{} embeddings preserve information about the task and the correctness of an execution while minimizing idiosyncratic information about the system, model family, and harness that produced it.

We then proposed using \trubric{} embeddings to induce representations of agentic systems for task-efficient evaluation.
The resulting geometry supports accurate benchmark score prediction from as little as a single executed task and, combined with an item-response model, is never worse than either constituent at any task budget when the full set of reference systems is available.

The combined estimator reduces the cost of reaching an MAE below 0.10 on SWE-bench Verified from \$10--\$40 to \$2 per system.
These results are relatively robust to the major design choices: the text encoder, the number of rubric fields, and the task selection method.
Importantly, these results support the claim that extracting more than binary correctness from cached executions can improve the efficiency of agentic evaluation as the number of systems and benchmarks continues to grow -- providing a route toward improved evaluation coverage for a given system.

\subsection{Limitations}

We evaluated \trubric{} along the five attributes relevant to efficient evaluation.
We make no claims about the usefulness of the representation for other tasks one might consider with traces, such as change detection \citep{bridgeford2025perspective}, online monitoring \citep{helm2026control}, early stopping \citep{ruan2026doomed, wang2026failfast}, or steering \citep{rahn2024entropic}.
We leave the study of \trubric{}-like representations for these tasks to future work.

As mentioned above, we did not attempt to optimize the rubric.
The rubric size experiment of Section~\ref{sec:results} demonstrates that favorable small rubrics exist.
Developing validated procedures for selecting or learning rubric fields, such as parameterizing meta-instructions, is an open problem that we expect to have a small to medium effect on accuracy.
We also note that \trubric{}'s fields can be reconfigured to amplify or dampen arbitrary dimensions of variation in the input data.

In our analysis we only consider DeepSeek-V4-Flash (accessed via OpenRouter) as the rubric generator and trace summarizer.
We had intended to add other frontier models as part of our ablation study, but a provider's API service banned our account and accused us of attempting to distill reasoning traces.
We include this note as a warning: summarizing traces at the scale required for trace-based evaluation can resemble distillation to automated enforcement systems and may result in account bans.
In any case, we leave the sensitivity analysis of the two roles to future work.

Finally, we only consider highly curated, paired settings where every reference system has executed (nearly) every task of a single benchmark.
We do not consider references with sparse or disjoint task coverage, nor signal shared across benchmarks.
The PKPS formulation we use throughout supports both -- Eq. \eqref{method:pkps} is defined for unpaired task sets -- and we expect less curated, larger reference collections to further reduce evaluation cost.

\subsection*{Acknowledgments}
We would like to thank Brandon Duderstadt, Youngser Park, and Carey Priebe for their feedback throughout the development of this manuscript.
We gratefully acknowledge funding from the Defense Advanced Research Projects Agency (DARPA) Artificial Intelligence Quantified (AIQ) award number HR00112520026.


\subsection*{AI use statement}
In this work, we used generative AI tools for implementing methods and simple data processing.
We also used AI in the creation of artifacts (e.g., prompting a coding agent to use Matplotlib to create figures) and to draft and edit parts of the manuscript with heavy user guidance.
We have reviewed all AI-assisted work. 
For example, generated code was verified and tested for correctness by all authors and
generated text was heavily edited.
We take responsibility for the final content of this work,
including text, claims or artifacts produced with the aid of generative AI.



\bibliography{references}

@inproceedings{jimenez2024swebench,
  title={{SWE}-bench: Can Language Models Resolve Real-World GitHub Issues?},
  author={Jimenez, Carlos E. and Yang, John and Wettig, Alexander and Yao, Shunyu and Pei, Kexin and Press, Ofir and Narasimhan, Karthik},
  booktitle={International Conference on Learning Representations (ICLR)},
  year={2024},
  note={arXiv:2310.06770}
}

@misc{openai2024verified,
  title={Introducing {SWE}-bench Verified},
  author={{OpenAI}},
  year={2024},
  howpublished={\url{https://openai.com/index/introducing-swe-bench-verified/}}
}

@article{tbench2026,
  title={Terminal-Bench: Benchmarking Agents on Hard, Realistic Tasks in Command Line Interfaces},
  author={Merrill, Mike A. and Shaw, Alexander G. and Carlini, Nicholas and others},
  journal={arXiv preprint arXiv:2601.11868},
  year={2026},
  doi={10.48550/arXiv.2601.11868}
}

@inproceedings{kapoor2025hal,
  title={Holistic Agent Leaderboard: The Missing Infrastructure for {AI} Agent Evaluation},
  author={Kapoor, Sayash and Stroebl, Benedikt and Kirgis, Peter and Nadgir, Nitya and Siegel, Zachary S. and others},
  booktitle={International Conference on Learning Representations (ICLR)},
  year={2026},
  note={arXiv:2510.11977}
}

@inproceedings{yang2024sweagent,
  title={{SWE}-agent: Agent-Computer Interfaces Enable Automated Software Engineering},
  author={Yang, John and Jimenez, Carlos E. and Wettig, Alexander and Lieret, Kilian and Yao, Shunyu and Narasimhan, Karthik and Press, Ofir},
  booktitle={Advances in Neural Information Processing Systems (NeurIPS)},
  year={2024},
  note={arXiv:2405.15793}
}

@inproceedings{wang2025openhands,
  title={{OpenHands}: An Open Platform for {AI} Software Developers as Generalist Agents},
  author={Wang, Xingyao and others},
  booktitle={International Conference on Learning Representations (ICLR)},
  year={2025},
  note={arXiv:2407.16741}
}

@article{mehtiyev2026beyond,
  title={Beyond Resolution Rates: Behavioral Drivers of Coding Agent Success and Failure},
  author={Mehtiyev, Tural and Assun\c{c}\~{a}o, Wesley},
  journal={arXiv preprint arXiv:2604.02547},
  year={2026},
  doi={10.48550/arXiv.2604.02547}
}

@article{shu2026traceprobe,
  title={What Resolve Rate Hides: Trajectory Structure Diagnostics for Coding Agents},
  author={Shu, Rui and Chong, Chun Yong and Zhou, Xin and Peng, Yun and Wu, Zihan and Han, Xu and Zhuang, Zeyang and Yuan, Guowen and Wang, Yuan},
  journal={arXiv preprint arXiv:2607.06184},
  year={2026},
  doi={10.48550/arXiv.2607.06184}
}

@inproceedings{cemri2025mast,
  title={Why Do Multi-Agent {LLM} Systems Fail?},
  author={Cemri, Mert and Pan, Melissa Z. and Yang, Shuyi and Agrawal, Lakshya A. and Chopra, Bhavya and Tiwari, Rishabh and Keutzer, Kurt and Parameswaran, Aditya and Klein, Dan and Ramchandran, Kannan and Zaharia, Matei and Gonzalez, Joseph E. and Stoica, Ion},
  booktitle={Advances in Neural Information Processing Systems (NeurIPS)},
  year={2025},
  note={arXiv:2503.13657}
}

@inproceedings{zhuge2025agentjudge,
  title={Agent-as-a-Judge: Evaluate Agents with Agents},
  author={Zhuge, Mingchen and Zhao, Changsheng and Ashley, Dylan R. and Wang, Wenyi and Khizbullin, Dmitrii and Xiong, Yunyang and Liu, Zechun and Chang, Ernie and Krishnamoorthi, Raghuraman and Tian, Yuandong and Shi, Yangyang and Chandra, Vikas and Schmidhuber, J{\"u}rgen},
  booktitle={International Conference on Machine Learning (ICML)},
  year={2025},
  note={arXiv:2410.10934}
}

@inproceedings{maiapolo2024tinybenchmarks,
  title={tiny{B}enchmarks: Evaluating {LLM}s with Fewer Examples},
  author={Maia Polo, Felipe and Weber, Lucas and Choshen, Leshem and Sun, Yuekai and Xu, Gongjun and Yurochkin, Mikhail},
  booktitle={International Conference on Machine Learning (ICML)},
  year={2024},
  note={arXiv:2402.14992}
}

@inproceedings{vivek2024anchor,
  title={Anchor Points: Benchmarking Models with Much Fewer Examples},
  author={Vivek, Rajan and Ethayarajh, Kawin and Yang, Diyi and Kiela, Douwe},
  booktitle={Conference of the European Chapter of the Association for Computational Linguistics (EACL)},
  year={2024},
  doi={10.18653/v1/2024.eacl-long.95},
  note={arXiv:2309.08638}
}

@inproceedings{kipnis2025metabench,
  title={{metabench} -- A Sparse Benchmark of Reasoning and Knowledge in Large Language Models},
  author={Kipnis, Alex and Voudouris, Konstantinos and Schulze Buschoff, Luca M. and Schulz, Eric},
  booktitle={International Conference on Learning Representations (ICLR)},
  year={2025},
  note={arXiv:2407.12844}
}

@inproceedings{truong2025amortized,
  title={Reliable and Efficient Amortized Model-based Evaluation},
  author={Truong, Sang T. and Tu, Yuheng and Liang, Percy and Li, Bo and Koyejo, Sanmi},
  booktitle={International Conference on Machine Learning (ICML)},
  year={2025},
  note={arXiv:2503.13335}
}

@inproceedings{ruan2024observational,
  title={Observational Scaling Laws and the Predictability of Language Model Performance},
  author={Ruan, Yangjun and Maddison, Chris J. and Hashimoto, Tatsunori},
  booktitle={Advances in Neural Information Processing Systems (NeurIPS)},
  year={2024},
  note={arXiv:2405.10938}
}

@inproceedings{zhuang2025embedllm,
  title={{EmbedLLM}: Learning Compact Representations of Large Language Models},
  author={Zhuang, Richard and Wu, Tianhao and Wen, Zhaojin and Li, Andrew and Jiao, Jiantao and Ramchandran, Kannan},
  booktitle={International Conference on Learning Representations (ICLR)},
  year={2025},
  note={arXiv:2410.02223}
}

@article{duderstadt2023data,
  title={Comparing Foundation Models using Data Kernels},
  author={Duderstadt, Brandon and Helm, Hayden S. and Priebe, Carey E.},
  journal={arXiv preprint arXiv:2305.05126},
  year={2023},
  doi={10.48550/arXiv.2305.05126}
}

@inproceedings{helm2025statistical,
  title={Statistical Inference on Black-Box Generative Models in the Data Kernel Perspective Space},
  author={Helm, Hayden and Acharyya, Aranyak and Park, Youngser and Duderstadt, Brandon and Priebe, Carey},
  booktitle={Findings of the Association for Computational Linguistics (ACL)},
  year={2025},
  pages={3955--3970},
  doi={10.18653/v1/2025.findings-acl.204},
  note={arXiv:2410.01106}
}

@inproceedings{helm2026query,
  title={Query-Efficient Model Evaluation Using Cached Responses},
  author={Helm, Hayden and Johnson, Ben and Priebe, Carey},
  booktitle={International Conference on Machine Learning (ICML)},
  year={2026},
  note={arXiv:2605.07096}
}

@article{acharyya2024consistent,
  title={Consistent Estimation of Generative Model Representations in the Data Kernel Perspective Space},
  author={Acharyya, Aranyak and Trosset, Michael W. and Priebe, Carey E. and Helm, Hayden S.},
  journal={arXiv preprint arXiv:2409.17308},
  year={2024},
  doi={10.48550/arXiv.2409.17308}
}

@inproceedings{reimers2019sbert,
  title={Sentence-{BERT}: Sentence Embeddings using Siamese {BERT}-Networks},
  author={Reimers, Nils and Gurevych, Iryna},
  booktitle={Conference on Empirical Methods in Natural Language Processing and International Joint Conference on Natural Language Processing (EMNLP-IJCNLP)},
  year={2019},
  pages={3982--3992},
  doi={10.18653/v1/D19-1410},
  note={arXiv:1908.10084}
}

@inproceedings{su2023instructor,
  title={One Embedder, Any Task: Instruction-Finetuned Text Embeddings},
  author={Su, Hongjin and Shi, Weijia and Kasai, Jungo and Wang, Yizhong and Hu, Yushi and Ostendorf, Mari and Yih, Wen-tau and Smith, Noah A. and Zettlemoyer, Luke and Yu, Tao},
  booktitle={Findings of the Association for Computational Linguistics (ACL)},
  year={2023},
  pages={1102--1121},
  doi={10.18653/v1/2023.findings-acl.71},
  note={arXiv:2212.09741}
}

@inproceedings{peng2024inbedder,
  title={Answer is All You Need: Instruction-following Text Embedding via Answering the Question},
  author={Peng, Letian and Zhang, Yuwei and Wang, Zilong and Srinivasa, Jayanth and Liu, Gaowen and Wang, Zihan and Shang, Jingbo},
  booktitle={Annual Meeting of the Association for Computational Linguistics (ACL)},
  year={2024},
  pages={459--477},
  doi={10.18653/v1/2024.acl-long.27},
  note={arXiv:2402.09642}
}

@inproceedings{benara2024qaemb,
  title={Crafting Interpretable Embeddings for Language Neuroscience by Asking {LLM}s Questions},
  author={Benara, Vinamra and Singh, Chandan and Morris, John X. and Antonello, Richard J. and Stoica, Ion and Huth, Alexander G. and Gao, Jianfeng},
  booktitle={Advances in Neural Information Processing Systems (NeurIPS)},
  year={2024},
  note={arXiv:2405.16714}
}

@inproceedings{ravfogel2024description,
  title={Description-Based Text Similarity},
  author={Ravfogel, Shauli and Pyatkin, Valentina and Cohen, Amir DN and Manevich, Avshalom and Goldberg, Yoav},
  booktitle={Conference on Language Modeling (COLM)},
  year={2024},
  note={arXiv:2305.12517}
}

@inproceedings{sun2025idiosyncrasies,
  title={Idiosyncrasies in Large Language Models},
  author={Sun, Mingjie and Yin, Yida and Xu, Zhiqiu and Kolter, J. Zico and Liu, Zhuang},
  booktitle={International Conference on Machine Learning (ICML)},
  year={2025},
  note={arXiv:2502.12150}
}

@article{oderinwale2026trajectories,
  title={Agent Trajectories as Programs: Fingerprinting and Programming Coding-Agent Behavior},
  author={Oderinwale, Hamidah},
  journal={arXiv preprint arXiv:2606.16988},
  year={2026},
  doi={10.48550/arXiv.2606.16988}
}

@article{nussbaum2025trainingsparsemixtureexperts,
  title={Training Sparse Mixture of Experts Text Embedding Models},
  author={Nussbaum, Zach and Duderstadt, Brandon},
  journal={arXiv preprint arXiv:2502.07972},
  year={2025},
  doi={10.48550/arXiv.2502.07972}
}

@incollection{birnbaum1968some,
  title={Some Latent Trait Models and Their Use in Inferring an Examinee's Ability},
  author={Birnbaum, Allan},
  booktitle={Statistical Theories of Mental Test Scores},
  editor={Lord, Frederic M. and Novick, Melvin R.},
  pages={397--479},
  year={1968},
  publisher={Addison-Wesley}
}

@article{deepseek2026v4,
  title={{DeepSeek-V4}: Towards Highly Efficient Million-Token Context Intelligence},
  author={{DeepSeek-AI}},
  journal={arXiv preprint arXiv:2606.19348},
  year={2026},
  doi={10.48550/arXiv.2606.19348}
}

@article{bowyer2026efficient,
  title={Efficient Benchmarking Is Just Feature Selection and Multiple Regression},
  author={Bowyer, Sam and Locatelli, Acyr and Cao, Kris},
  journal={arXiv preprint arXiv:2605.25773},
  year={2026},
  doi={10.48550/arXiv.2605.25773}
}

@article{torgerson1952mds,
  title={Multidimensional scaling: {I}. Theory and method},
  author={Torgerson, Warren S},
  journal={Psychometrika},
  volume={17},
  number={4},
  pages={401--419},
  year={1952},
  doi={10.1007/BF02288916}
}

@article{neelakantan2022text,
  title={Text and code embeddings by contrastive pre-training},
  author={Neelakantan, Arvind and Xu, Tao and Puri, Raul and Radford, Alec and Han, Jesse Michael and Tworek, Jerry and Yuan, Qiming and Tezak, Nikolas and Kim, Jong Wook and Hallacy, Chris and others},
  journal={arXiv preprint arXiv:2201.10005},
  year={2022},
  doi={10.48550/arXiv.2201.10005}
}

@article{nussbaum2024nomic,
  title={{Nomic Embed}: Training a Reproducible Long Context Text Embedder},
  author={Nussbaum, Zach and Morris, John X. and Duderstadt, Brandon and Mulyar, Andriy},
  journal={Transactions on Machine Learning Research},
  issn={2835-8856},
  year={2025},
  url={https://openreview.net/forum?id=IPmzyQSiQE}
}

@inproceedings{xiao2023cpack,
  title={{C-Pack}: Packed Resources for General {Chinese} Embeddings},
  author={Xiao, Shitao and Liu, Zheng and Zhang, Peitian and Muennighoff, Niklas and Lian, Defu and Nie, Jian-Yun},
  booktitle={Proceedings of the 47th International ACM SIGIR Conference on Research and Development in Information Retrieval},
  pages={641--649},
  year={2024},
  doi={10.1145/3626772.3657878}
}

@article{li2023gte,
  title={Towards general text embeddings with multi-stage contrastive learning},
  author={Li, Zehan and Zhang, Xin and Zhang, Yanzhao and Long, Dingkun and Xie, Pengjun and Zhang, Meishan},
  journal={arXiv preprint arXiv:2308.03281},
  year={2023},
  doi={10.48550/arXiv.2308.03281}
}

@article{wang2022e5,
  title={Text embeddings by weakly-supervised contrastive pre-training},
  author={Wang, Liang and Yang, Nan and Huang, Xiaolong and Jiao, Binxing and Yang, Linjun and Jiang, Daxin and Majumder, Rangan and Wei, Furu},
  journal={arXiv preprint arXiv:2212.03533},
  year={2022},
  doi={10.48550/arXiv.2212.03533}
}

@article{bridgeford2025perspective,
  title={Detecting Perspective Shifts in Multi-agent Systems},
  author={Bridgeford, Eric and Helm, Hayden},
  journal={arXiv preprint arXiv:2512.05013},
  year={2025},
  doi={10.48550/arXiv.2512.05013}
}

@article{helm2026control,
  title={Control Charts for Multi-agent Systems},
  author={Helm, Hayden and Priebe, Carey and Duderstadt, Brandon},
  journal={arXiv preprint arXiv:2605.11135},
  year={2026},
  doi={10.48550/arXiv.2605.11135}
}

@article{ruan2026doomed,
  title={Doomed from the Start: Early Abort of {LLM} Agent Episodes via a Recall-Controlled Probe Cascade},
  author={Ruan, Kai and Huang, Zihe and Zhou, Ziqi and Wei, Qianshan and Lin, Jinghao and Wang, Xuan and Sun, Hao},
  journal={arXiv preprint arXiv:2607.06503},
  year={2026},
  doi={10.48550/arXiv.2607.06503}
}

@article{wang2026failfast,
  title={Fail-Fast, Restart-Smart: Early Failure Prediction and Restart for {SWE} Agentic Tasks},
  author={Wang, Chenyu and Lyu, Yunbo and He, Junda and Yang, Zhou and Zhong, Chenxing and Harel, Yaniv and Lo, David},
  journal={arXiv preprint arXiv:2608.03222},
  year={2026},
  doi={10.48550/arXiv.2608.03222}
}

@article{rahn2024entropic,
  title={Controlling Large Language Model Agents with Entropic Activation Steering},
  author={Rahn, Nate and D'Oro, Pierluca and Bellemare, Marc G.},
  journal={arXiv preprint arXiv:2406.00244},
  year={2024},
  doi={10.48550/arXiv.2406.00244}
}

@misc{helm2026benchmarkpredictionsparseunpaired,
      title={Benchmark Prediction with Sparse and Unpaired Cached Responses}, 
      author={Hayden Helm and Avijit Ghosh and Ben Johnson and Carey Priebe},
      year={2026},
      eprint={2611.03614},
      archivePrefix={arXiv},
      primaryClass={cs.LG},
      url={https://arxiv.org/abs/2611.03614}, 
      doi={10.48550/arXiv.2611.03614}
}
\bibliographystyle{iclr2027/iclr2027_conference}

\appendix

\section{Trace representation details}
\label{app:traces}

\subsection{Example task, trace, rubrics, and extractions}
\label{app:example}

We illustrate the \trubric{} pipeline on SWE-bench Verified instance \texttt{django\_\_django-11099}.

\paragraph{Task.} The problem statement, as provided to the rubric-instantiation LLM:
\begin{quote}\small
UsernameValidator allows trailing newline in usernames \\ Description \\ 	 \\ ASCIIUsernameValidator and UnicodeUsernameValidator use the regex  \\ r'\textasciicircum{}[\textbackslash{}w.@+-]+\$' \\ The intent is to only allow alphanumeric characters as well as ., @, +, and -. However, a little known quirk of Python regexes is that \$ will also match a trailing newline. Therefore, the user name validators will accept usernames which end with a newline. You can avoid this behavior by instead using \textbackslash{}A and \textbackslash{}Z to terminate regexes. For example, the validator regex could be changed to \\ r'\textbackslash{}A[\textbackslash{}w.@+-]+\textbackslash{}Z' \\ in order to reject usernames that end with a newline. \\ I am not sure how to officially post a patch, but the required change is trivial - using the regex above in the two validators in contrib.auth.validators.
\end{quote}

\paragraph{Instantiated \trubric{}.} The six task-specific field descriptions generated for this instance:
\begin{description}[itemsep=1pt, topsep=2pt]\small
\item[understanding] Look for the agent to identify that the issue lies in the regex anchoring with \$ in ASCIIUsernameValidator and UnicodeUsernameValidator, specifically that \$ matches a trailing newline in Python, and that changing to \textbackslash{}A and \textbackslash{}Z is the proposed fix in django/contrib/auth/validators.py.
\item[localization] Check that the agent correctly locates the two validator classes (ASCIIUsernameValidator and UnicodeUsernameValidator) in django/contrib/auth/validators.py and references the existing regex r'\textasciicircum{}[\textbackslash{}w.@+-]+\$' in both.
\item[reproduction] Look for the agent to write or run a test that confirms a username ending with a newline (e.g., 'validuser\textbackslash{}n') is currently accepted by the validators, possibly by using Django's validate\_username or directly instantiating the validators.
\item[editing] Verify that the agent modifies the regex pattern in both validator classes from r'\textasciicircum{}[\textbackslash{}w.@+-]+\$' to r'\textbackslash{}A[\textbackslash{}w.@+-]+\textbackslash{}Z', ensuring no other changes are made to the file.
\item[verification] The agent should run the relevant test suite (e.g., tests/auth\_tests/test\_validators.py or full Django auth tests) to confirm that usernames with trailing newlines are now rejected, and that existing valid usernames still pass.
\item[final state] The final state should show django/contrib/auth/validators.py with both regex patterns updated, and no other unintended modifications, with tests confirming the newline rejection behavior.
\end{description}

\paragraph{Generic rubric.} For contrast, the generic baseline uses the same six fields with fixed, task-agnostic descriptions shared by every task:
\begin{description}[itemsep=1pt, topsep=2pt]\small
\item[understanding] how the agent oriented itself before acting
\item[localization] how it searched/navigated to find the relevant code
\item[reproduction] whether/how it reproduced the issue before fixing
\item[editing] what it changed (files, nature and size of the edit)
\item[verification] whether it re-ran tests or a repro script after editing, and the OBSERVED outcome (passing/failing/errors)
\item[final state] how the run ended (clean finish, step/cost limit, submitted unverified)
\end{description}

\paragraph{Trace.} An excerpt of System A's rendered trace on this task, as provided to the summarization LLM.
The full rendered trace is 98{,}806 characters; its beginning and end are shown.
\begin{quote}\scriptsize\begin{verbatim}
swe_main
First, we should locate the files that contain the `ASCIIUsernameValidator` and
`UnicodeUsernameValidator` classes to understand where the regex patterns are
defined. Since the issue mentions that these validators are in
`contrib.auth.validators`, we can start by searching for files named
`validators.py` within the `contrib/auth` directory.


action: find_file "validators.py" contrib/auth

Directory contrib/auth not found

First, we should locate the files that contain the `ASCIIUsernameValidator` and
`UnicodeUsernameValidator` classes to understand where the regex patterns are
defined. [...]

```
find_file "validators.py" contrib/auth
```
{"open_file": "n/a", "working_dir": "/django__django"}

[... roughly 97,000 characters omitted ...]

action: python regex_test.py

primary
role: user
ASCIIUsernameValidator rejected the username.
UnicodeUsernameValidator rejected the username.

(Open file: /django__django/regex_test.py)
(Current directory: /django__django)
bash-$
primary
role: assistant
The output indicates that both the `ASCIIUsernameValidator` and
`UnicodeUsernameValidator` regex patterns correctly rejected the username with
a trailing newline. This confirms that the changes made to the regex patterns
are effective.

Since the issue has been addressed and the fix has been verified with a test
script, we can now submit our changes to the code base.

```
submit
```
action: submit
\end{verbatim}\end{quote}

\paragraph{Extractions.} Structured summaries of two systems' traces on this task, produced against the rubric above. The summaries describe execution behavior without reference to the harness or underlying model that produced the trace.

\subparagraph{System A.}
\begin{description}[itemsep=1pt, topsep=2pt]\small
\item[understanding] The agent identified that the issue stems from Python's regex behavior where \$ matches a trailing newline, and that changing the anchors to \textbackslash{}A and \textbackslash{}Z in both validators would fix it. It explicitly mentioned this during the initial inspection of the validator file.
\item[localization] The agent searched for validators.py, found django/contrib/auth/validators.py, opened it, and located both ASCIIUsernameValidator and UnicodeUsernameValidator classes. It acknowledged the existing regex pattern r'\textasciicircum{}[\textbackslash{}w.@+-]+\$' in both classes before making changes.
\item[reproduction] The agent attempted to reproduce the bug by creating test\_validators.py and simple\_test\_validators.py, but these scripts failed due to missing Django settings configuration. It then created regex\_test.py to directly test the regex patterns, which ran successfully and confirmed rejection of a trailing newline after the fix, though it did not demonstrate acceptance before the fix.
\item[editing] The agent edited lines 10 and 20 in django/contrib/auth/validators.py, changing the regex from r'\textasciicircum{}[\textbackslash{}w.@+-]+\$' to r'\textbackslash{}A[\textbackslash{}w.@+-]+\textbackslash{}Z' in both ASCIIUsernameValidator and UnicodeUsernameValidator. The diff shows only these two regex changes to the file, with no other modifications.
\item[verification] The agent ran the custom regex\_test.py script, which output that both ASCII and Unicode validators rejected a username ending with a newline. It did not run the existing Django test suite (e.g., tests/auth\_tests/test\_validators.py) to verify the fix against formal tests.
\item[final state] The final state includes django/contrib/auth/validators.py with both regex patterns updated to r'\textbackslash{}A[\textbackslash{}w.@+-]+\textbackslash{}Z'. Three new files (test\_validators.py, simple\_test\_validators.py, regex\_test.py) were created, but no other existing files were modified. The agent submitted these changes.
\end{description}

\subparagraph{System B.}
\begin{description}[itemsep=1pt, topsep=2pt]\small
\item[understanding] The agent identified that the issue stems from Python's regex behavior where \$ matches a trailing newline, causing ASCIIUsernameValidator and UnicodeUsernameValidator to incorrectly accept usernames ending with newlines. It understood that replacing \textasciicircum{} and \$ with \textbackslash{}A and \textbackslash{}Z would fix the anchoring issue, as described in the issue for django/contrib/auth/validators.py.
\item[localization] The agent located the two validator classes, ASCIIUsernameValidator and UnicodeUsernameValidator, in django/contrib/auth/validators.py. It confirmed both classes use the existing regex pattern r'\textasciicircum{}[\textbackslash{}w.@+-]+\$' and identified these as the exact locations requiring modification.
\item[reproduction] The agent verified the issue by running a quick test comparing the old and new regex patterns against usernames with and without trailing newlines. The test confirmed that the old pattern matched usernames ending with a newline, while the new pattern correctly rejected them, and valid usernames still passed.
\item[editing] The agent modified the regex pattern in both validator classes from r'\textasciicircum{}[\textbackslash{}w.@+-]+\$' to r'\textbackslash{}A[\textbackslash{}w.@+-]+\textbackslash{}Z' in django/contrib/auth/validators.py. The diff shows only these two regex changes, with no other modifications made to the file.
\item[verification] The agent ran the auth test suite which passed all 494 tests, including the tests in auth\_tests/test\_validators.py for UsernameValidatorsTests. It also ran a targeted test confirming that usernames with trailing newlines are now rejected by both validators while valid usernames still pass.
\item[final state] The final state shows django/contrib/auth/validators.py with both regex patterns updated to r'\textbackslash{}A[\textbackslash{}w.@+-]+\textbackslash{}Z' in ASCIIUsernameValidator and UnicodeUsernameValidator. No other files were modified, and tests confirm the newline rejection behavior works correctly.
\end{description}

\subsection{Linear classification tasks}
\label{app:probes}

For each combination of text encoder and trace representation, we assemble one embedding per trace for the executions of the 107 SWE-bench Verified systems on a fixed panel of 20 tasks (2{,}129 traces; a small number of executions are missing from the public archive).
Rubric-based representations are the concatenated per-field embeddings of Section~\ref{sec:method-trubric}; raw-trace representations embed head and tail slices of the rendered log as two blocks.
Each block is $\ell_{2}$-normalized and blocks are weighted equally.

Probes are linear classifiers trained with class-balanced squared loss and a ridge penalty; the penalty is selected along a closed-form regularization path using grouped cross-validation within the training folds.
Evaluation is five-fold grouped cross-validation with seeded, size-balanced group allocation: for the system-level attributes (iii)--(v) the groups are tasks, so probes are evaluated on tasks unseen during training, and for (i)--(ii) the groups are model families, so probes are evaluated on systems from families unseen during training.
We report balanced accuracy averaged over folds; chance balanced accuracy for a $C$-class attribute is $1/C$.
Figure~\ref{fig:radar} plots $(\mathrm{accuracy} - \mathrm{chance})/(1 - \mathrm{chance})$ for attributes (i)--(ii) and its complement for (iii)--(v), so that better-for-evaluation is always outward.
Table~\ref{tab:probes} reports the balanced accuracies underlying Figure~\ref{fig:radar}.

\begin{table}[t]
\centering
\caption{Held-out balanced accuracy of linear probes for the five trace attributes of Section~\ref{sec:method-trubric}, for each text encoder and trace representation; these values underlie Figure~\ref{fig:radar}. For evaluation-relevant attributes (i)--(ii), higher is better; for system-metadata attributes (iii)--(v), lower is better. The final row reports chance balanced accuracy.}
\label{tab:probes}
\tiny
\begin{tabular}{llccccc}
\toprule
encoder & representation & {(i) task} & {(ii) correctness} & {(iii) system} & {(iv) family} & {(v) harness} \\
\midrule
\texttt{text-embedding-3-small} & raw trace & 1.00 & 0.82 & 0.89 & 0.97 & 0.96 \\
 & generic & 0.99 & 0.84 & 0.61 & 0.64 & 0.76 \\
 & \trubric{} & 0.99 & 0.84 & 0.20 & 0.36 & 0.29 \\
\addlinespace[2pt]
\texttt{nomic-embed-text-v1.5} & raw trace & 0.98 & 0.78 & 0.84 & 0.90 & 0.91 \\
 & generic & 0.99 & 0.86 & 0.56 & 0.63 & 0.70 \\
 & \trubric{} & 0.99 & 0.84 & 0.16 & 0.30 & 0.22 \\
\addlinespace[2pt]
\texttt{bge-large-en-v1.5} & raw trace & 0.95 & 0.78 & 0.73 & 0.79 & 0.86 \\
 & generic & 0.99 & 0.84 & 0.58 & 0.64 & 0.73 \\
 & \trubric{} & 0.99 & 0.85 & 0.16 & 0.33 & 0.23 \\
\addlinespace[2pt]
\texttt{gte-large} & raw trace & 0.95 & 0.77 & 0.77 & 0.83 & 0.90 \\
 & generic & 0.99 & 0.85 & 0.56 & 0.59 & 0.70 \\
 & \trubric{} & 0.99 & 0.84 & 0.17 & 0.33 & 0.23 \\
\addlinespace[2pt]
\texttt{e5-large-v2} & raw trace & 0.96 & 0.77 & 0.79 & 0.86 & 0.90 \\
 & generic & 0.99 & 0.85 & 0.62 & 0.66 & 0.77 \\
 & \trubric{} & 0.99 & 0.85 & 0.21 & 0.37 & 0.29 \\
\addlinespace[2pt]
\texttt{all-mpnet-base-v2} & raw trace & 0.93 & 0.77 & 0.68 & 0.76 & 0.82 \\
 & generic & 0.99 & 0.84 & 0.43 & 0.52 & 0.57 \\
 & \trubric{} & 0.99 & 0.83 & 0.11 & 0.29 & 0.16 \\
\addlinespace[2pt]
\texttt{all-MiniLM-L6-v2} & raw trace & 0.91 & 0.75 & 0.70 & 0.75 & 0.86 \\
 & generic & 0.99 & 0.84 & 0.45 & 0.53 & 0.59 \\
 & \trubric{} & 0.99 & 0.82 & 0.11 & 0.31 & 0.16 \\
\addlinespace[2pt]
\midrule
\multicolumn{2}{l}{chance} & 0.05 & 0.50 & 0.01 & 0.12 & 0.02 \\
\bottomrule
\end{tabular}
\end{table}

\subsection{Prompts}
\label{app:prompts}

The rubric-instantiation prompt (\texttt{\{problem\}} is the task description):
\begin{quote}\scriptsize\begin{verbatim}
Here is a repository issue that a software-engineering agent will be asked to fix:

{problem}

Write an instance-specific rubric for judging an agent's trace on THIS issue:
for each section below, one sentence saying what specifically to look for in
this instance (name the likely relevant modules/files/tests if inferable).

Sections: understanding, localization, reproduction, editing,
verification, final_state.
Respond with ONLY a JSON object with those keys, each a one-sentence string.
\end{verbatim}\end{quote}

The extraction prompt (\texttt{\{rubric\}} is the instantiated \trubric{} for the task, or the fixed generic rubric below for the generic baseline; \texttt{\{keys\}} the field names; the rendered trace is appended after the prompt):
\begin{quote}\scriptsize\begin{verbatim}
Below is the (truncated) execution trace of a software-engineering agent.
Extract a factual description of the agent's behavior for EACH of the following
sections. 30-60 words per section. Describe behavior only -- do not restate the
problem. If a section did not occur, write exactly what is missing (e.g. "never
ran any test after editing").

{rubric}

Respond with ONLY a JSON object with keys {keys}, each a string.
\end{verbatim}\end{quote}

The generic rubric:
\begin{quote}\scriptsize\begin{verbatim}
- understanding: how the agent oriented itself before acting
- localization: how it searched/navigated to find the relevant code
- reproduction: whether/how it reproduced the issue before fixing
- editing: what it changed (files, nature and size of the edit)
- verification: whether it re-ran tests or a repro script after editing,
  and the OBSERVED outcome (passing/failing/errors)
- final_state: how the run ended (clean finish, step/cost limit,
  submitted unverified)
\end{verbatim}\end{quote}

\subsection{Rubric field list}
\label{app:fields}

The rubric-size experiment of Figure~\ref{fig:sensitivity} draws field subsets from a bank of 32 fields.
The bank consists of the six canonical fields used throughout the paper -- understanding, localization, reproduction, editing, verification, and final state -- together with 26 additional fields: requirements, causal diagnosis, code context, planning, test design, edge cases, regression scope, feedback response, tool recovery, patch review, exploration breadth, hypothesis testing, search strategy, dependency handling, error interpretation, incrementalism, backtracking, code reuse, instruction adherence, efficiency, state tracking, assumption checking, output quality, risk management, domain knowledge, and completion confidence.

\section{Efficient benchmarking details}
\label{app:baselines}

\subsection{Score-only baselines}
\label{app:score-baselines}

\paragraph{Sample score.} The mean of the target's $m$ observed trace-level scores, $\hat{y}_{\mathrm{sample}} = \frac{1}{m} \sum_{t \in T_{0}} s(f_{0}(t))$.

\paragraph{Population mean.} The mean benchmark score of the reference systems, $\hat{y}_{\mathrm{pop}} = \frac{1}{n} \sum_{i=1}^{n} y_{i}$, which ignores the target's executions entirely.

\paragraph{Item-response model.}
We fit the two-parameter logistic model $\mathbb{P}(s(x_{ij}) = 1) = \mathrm{sigmoid}\left(a_{j} (\theta_{i} - b_{j})\right)$ to the reference score matrix by maximum a posteriori estimation, with standard normal priors on the abilities $\theta_{i}$ and difficulties $b_{j}$ and a $\mathcal{N}(0, 1/10)$ prior on the log-discriminations $\log a_{j}$ (centering the slopes at the Rasch value of one), optimized with L-BFGS.
Because the benchmark score aggregates tasks beyond the $M$-task panel (all 500 tasks for SWE-bench Verified; all tasks and trials for Terminal-Bench 2.0), we additionally fit a monotone link from ability to benchmark score, $y \approx \mathrm{sigmoid}(c_{1} \theta + c_{0})$, by least squares on the reference systems.
For a target, we form a posterior over ability on a dense grid -- the prior is a normal with the mean and standard deviation of the reference abilities -- update it with the likelihood of the target's $m$ observed trace-level scores, and report the posterior mean of the fitted ability-to-score link as $\hat{y}_{\mathrm{IRT}}$.

\paragraph{Adaptive task selection.}
Given the fitted item-response model, tasks are selected greedily per target: at each step, the next task is the one with the highest expected Fisher information $a_{j}^{2} \, p_{j}(\theta)(1 - p_{j}(\theta))$ under the current ability posterior, the outcome is observed, and the posterior is updated.
Random task selection instead draws $T_{0}$ uniformly; reported random-selection results average over 50 shared draws.

\subsection{Trace-geometry baselines}
\label{app:geom-baselines}

The raw-trace and generic-rubric estimators replace the \trubric{} representation $\phi$ in the pipeline of Section~\ref{sec:method-pkps} and are otherwise identical.
The raw-trace representation embeds the first and last portions of the rendered execution log as two blocks with the same encoder $g$; representations are centered by the task-level median across systems and $\ell_{2}$-normalized.
The generic-rubric representation follows the two-stage construction of Section~\ref{sec:method-trubric} with the fixed field descriptions of Appendix~\ref{app:prompts} in place of task-specific rubrics.

\subsection{Hyperparameters and cross-validation}
\label{app:hyper}

For each task draw, hyperparameters are selected by treating each reference system as a pseudo-target: its score is predicted from the remaining references (leave-one-out via the hat matrix for the ridge regressor), and the configuration minimizing the mean reference error is selected.
The candidate grids are the kernel bandwidth $\sigma^{2} \in \{\mathrm{med}/2^{k}\}_{k=0}^{8}$ where $\mathrm{med}$ is the median squared distance between task embeddings, the dimension $d \in \{8, 16\}$, and the ridge penalty $\in \{0.1, 1.0\}$.
The ensemble weight $\alpha$ is selected on a $101$-point grid over $[0, 1]$ by the same reference-pseudo-target procedure.

\section{Systems and tasks}
\label{app:data}

\subsection{SWE-bench Verified systems}
Table~\ref{tab:swe-systems} lists the 107 evaluated leaderboard submissions with their model-family tag (used by the leave-one-family-out protocol; \emph{--} indicates no tag could be derived from submission metadata) and official resolve rate $y$.

\begin{center}\scriptsize
\begin{longtable}{llc}
\caption{SWE-bench Verified systems.}\label{tab:swe-systems}\\
\toprule
system & family & $y$ \\
\midrule
\endfirsthead
\toprule
system & family & $y$ \\
\midrule
\endhead
\bottomrule
\endlastfoot
\texttt{20240402\_sweagent\_gpt4} & gpt & 0.224 \\
\texttt{20240612\_MASAI\_gpt4o} & gpt & 0.326 \\
\texttt{20240620\_sweagent\_claude3.5sonnet} & claude & 0.336 \\
\texttt{20240721\_amazon-q-developer-agent-20240719-dev} & -- & 0.388 \\
\texttt{20240820\_epam-ai-run-gpt-4o} & gpt & 0.240 \\
\texttt{20240820\_honeycomb} & -- & 0.406 \\
\texttt{20240824\_gru} & -- & 0.452 \\
\texttt{20240918\_lingma-agent\_lingma-swe-gpt-72b} & gpt & 0.250 \\
\texttt{20240918\_lingma-agent\_lingma-swe-gpt-7b} & gpt & 0.102 \\
\texttt{20240920\_solver} & -- & 0.436 \\
\texttt{20240924\_solver} & -- & 0.454 \\
\texttt{20241001\_nfactorial} & -- & 0.258 \\
\texttt{20241002\_lingma-agent\_lingma-swe-gpt-72b} & gpt & 0.288 \\
\texttt{20241002\_lingma-agent\_lingma-swe-gpt-7b} & gpt & 0.182 \\
\texttt{20241016\_composio\_swekit} & claude & 0.406 \\
\texttt{20241016\_epam-ai-run-gpt-4o} & gpt & 0.270 \\
\texttt{20241022\_tools\_claude-3-5-haiku} & claude & 0.406 \\
\texttt{20241022\_tools\_claude-3-5-sonnet-updated} & claude & 0.490 \\
\texttt{20241023\_emergent} & -- & 0.466 \\
\texttt{20241025\_composio\_swekit} & -- & 0.486 \\
\texttt{20241028\_agentless-1.5\_gpt4o} & gpt & 0.388 \\
\texttt{20241028\_solver} & -- & 0.500 \\
\texttt{20241029\_OpenHands-CodeAct-2.1-sonnet-20241022} & claude & 0.530 \\
\texttt{20241029\_epam-ai-run-claude-3-5-sonnet} & claude & 0.396 \\
\texttt{20241030\_nfactorial} & -- & 0.416 \\
\texttt{20241105\_nfactorial} & -- & 0.492 \\
\texttt{20241106\_navie-2-gpt4o-sonnet} & -- & 0.472 \\
\texttt{20241108\_autocoderover-v2.0-claude-3-5-sonnet-20241022} & claude & 0.462 \\
\texttt{20241108\_devlo} & claude & 0.542 \\
\texttt{20241113\_nebius-search-open-weight-models-11-24} & -- & 0.406 \\
\texttt{20241120\_artemis\_agent} & -- & 0.320 \\
\texttt{20241125\_enginelabs} & claude & 0.518 \\
\texttt{20241125\_marscode-agent-dev} & -- & 0.500 \\
\texttt{20241202\_agentless-1.5\_claude-3.5-sonnet-20241022} & claude & 0.508 \\
\texttt{20241202\_amazon-q-developer-agent-20241202-dev} & -- & 0.550 \\
\texttt{20241208\_gru} & -- & 0.570 \\
\texttt{20241212\_epam-ai-run-claude-3-5-sonnet} & claude & 0.554 \\
\texttt{20241212\_google\_jules\_gemini\_2.0\_flash\_experimental} & gemini & 0.522 \\
\texttt{20241213\_devlo} & -- & 0.582 \\
\texttt{20241223\_emergent} & -- & 0.572 \\
\texttt{20250110\_blackboxai\_agent\_v1.1} & -- & 0.628 \\
\texttt{20250110\_learn\_by\_interact\_claude3.5} & claude & 0.602 \\
\texttt{20250117\_wandb\_programmer\_o1\_crosscheck5} & o1 & 0.646 \\
\texttt{20250118\_codeshellagent\_gemini\_2.0\_flash\_experimental} & gemini & 0.442 \\
\texttt{20250120\_Bracket} & -- & 0.532 \\
\texttt{20250122\_autocoderover-v2.1-claude-3-5-sonnet-20241022} & claude & 0.516 \\
\texttt{20250203\_openhands\_4x\_scaled} & -- & 0.608 \\
\texttt{20250206\_agentscope} & claude & 0.634 \\
\texttt{20250214\_agentless\_lite\_o3\_mini} & o3 & 0.424 \\
\texttt{20250224\_tools\_claude-3-7-sonnet} & claude & 0.632 \\
\texttt{20250226\_swerl\_llama3\_70b} & llama & 0.412 \\
\texttt{20250228\_epam-ai-run-claude-3-5-sonnet} & claude & 0.628 \\
\texttt{20250316\_augment\_agent\_v0} & -- & 0.654 \\
\texttt{20250405\_amazon-q-developer-agent-20250405-dev} & -- & 0.654 \\
\texttt{20250410\_cortexa} & -- & 0.582 \\
\texttt{20250415\_openhands} & -- & 0.658 \\
\texttt{20250430\_zencoder\_ai} & -- & 0.700 \\
\texttt{20250511\_sweagent\_lm\_32b} & -- & 0.402 \\
\texttt{20250514\_aime\_coder} & claude & 0.664 \\
\texttt{20250515\_Refact\_Agent} & -- & 0.704 \\
\texttt{20250516\_cortexa\_o3} & -- & 0.682 \\
\texttt{20250519\_devlo} & -- & 0.702 \\
\texttt{20250519\_trae} & -- & 0.706 \\
\texttt{20250520\_openhands\_devstral\_small} & devstral & 0.468 \\
\texttt{20250522\_sweagent\_claude-4-sonnet-20250514} & claude & 0.666 \\
\texttt{20250522\_tools\_claude-4-opus} & claude & 0.732 \\
\texttt{20250522\_tools\_claude-4-sonnet} & claude & 0.724 \\
\texttt{20250524\_openhands\_claude\_4\_sonnet} & claude & 0.704 \\
\texttt{20250527\_amazon.nova-premier-v1.0} & nova & 0.424 \\
\texttt{20250603\_Refact\_Agent\_claude-4-sonnet} & -- & 0.744 \\
\texttt{20250610\_augment\_agent\_v1} & -- & 0.704 \\
\texttt{20250611\_moatless\_claude-4-sonnet-20250514} & claude & 0.708 \\
\texttt{20250612\_trae} & -- & 0.752 \\
\texttt{20250616\_Skywork-SWE-32B} & qwen & 0.380 \\
\texttt{20250616\_Skywork-SWE-32B+TTS\_Bo8} & -- & 0.470 \\
\texttt{20250623\_warp} & -- & 0.710 \\
\texttt{20250629\_deepswerl\_r2eagent} & -- & 0.422 \\
\texttt{20250629\_deepswerl\_r2eagent\_tts} & o1 & 0.588 \\
\texttt{20250710\_bloop} & claude & 0.712 \\
\texttt{20250715\_qodo\_command} & -- & 0.712 \\
\texttt{20250716\_openhands\_kimi\_k2} & kimi & 0.654 \\
\texttt{20250720\_Lingxi-v1.5\_claude-4-sonnet-20250514} & claude & 0.746 \\
\texttt{20250725\_sweagent\_devstral\_small\_2507} & devstral & 0.380 \\
\texttt{20250728\_zai\_glm4-5} & -- & 0.642 \\
\texttt{20250731\_harness\_ai} & claude & 0.748 \\
\texttt{20250804\_codesweep\_sweagent\_kimi\_k2\_instruct} & kimi & 0.534 \\
\texttt{20250804\_epam-ai-run-claude-4-sonnet} & claude & 0.768 \\
\texttt{20250806\_SWE-Exp\_DeepSeek-V3} & -- & 0.420 \\
\texttt{20250807\_openhands\_gpt5} & gpt & 0.718 \\
\texttt{20250819\_ACoder} & -- & 0.764 \\
\texttt{20250901\_entroPO\_R2E\_QwenCoder30BA3B} & qwen & 0.522 \\
\texttt{20250901\_entroPO\_R2E\_QwenCoder30BA3B\_tts} & qwen & 0.604 \\
\texttt{20250901\_warp} & -- & 0.756 \\
\texttt{20250902\_atlassian-rovo-dev} & -- & 0.768 \\
\texttt{20250915\_JoyCode} & -- & 0.746 \\
\texttt{20250924\_artemis\_agent\_v2} & claude & 0.570 \\
\texttt{20250928\_trae\_doubao\_seed\_code} & doubao & 0.788 \\
\texttt{20250930\_zai\_glm4-6} & -- & 0.682 \\
\texttt{20251014\_Lingxi\_kimi\_k2} & kimi & 0.712 \\
\texttt{20251021\_SalesforceAIResearch\_SAGE\_bash\_only} & -- & 0.730 \\
\texttt{20251103\_SalesforceAIResearch\_SAGE\_OpenHands} & -- & 0.738 \\
\texttt{20251103\_sonar-foundation-agent\_claude-sonnet-4-5} & claude & 0.748 \\
\texttt{20251110\_frogboss-32b} & -- & 0.536 \\
\texttt{20251110\_frogmini-14b} & -- & 0.450 \\
\texttt{20251120\_livesweagent\_gemini-3-pro-preview} & gemini & 0.774 \\
\texttt{20251127\_openhands\_claude-opus-4-5} & claude & 0.776 \\
\texttt{20251205\_sonar-foundation-agent\_claude-opus-4-5} & claude & 0.792 \\
\end{longtable}
\end{center}

\clearpage
\subsection{Terminal-Bench 2.0 systems}
Table~\ref{tab:tb2-systems} lists the 65 evaluated submissions with their harness, underlying-model tags, and official score $y$ (mean over five trials). Truncated entries are spelled out below the table.

\begin{center}\scriptsize
\begin{longtable}{lllc}
\caption{Terminal-Bench 2.0 systems.}\label{tab:tb2-systems}\\
\toprule
system & harness & model tags & $y$ \\
\midrule
\endfirsthead
\toprule
system & harness & model tags & $y$ \\
\midrule
\endhead
\bottomrule
\endlastfoot
\texttt{Ante\_\_Gemini-3-Pro-Preview} & ante & gemini-3-pro-preview & 0.647 \\
\texttt{Ante\_\_Gemini-3.1-Pro-Preview} & ante & gemini-3-pro-preview & 0.742 \\
\texttt{BashAgent\_\_TermiGen-32B} & bash-agent & termigen-32b & 0.197 \\
\texttt{Capy\_\_GPT-5.5} & capy & gpt-5-5 & 0.834 \\
\texttt{ClaudeCode\_\_GLM-4.7} & claude-code & glm-4-7 & 0.333 \\
\texttt{CodeBrain-1.5\_\_GPT-5.3-Codex} & codebrain-1 & gpt-5-3-codex & 0.816 \\
\texttt{CodeBrain-1\_\_GPT-5.3-Codex} & codebrain-1 & gpt-5-3-codex & 0.703 \\
\texttt{CodeBrain-1\_\_Gemini-3-Pro-Preview} & codebrain-1 & gemini-3-pro-preview & 0.622 \\
\texttt{Codelia\_\_GPT-5.3-Codex} & codelia & gpt-5-3-codex & 0.757 \\
\texttt{Deep-Agents\_\_GPT-5.2-Codex} & deep-agents & gpt-5-2-codex & 0.667 \\
\texttt{Dirac\_\_Gemini-3-Flash-Preview} & dirac & gemini-3-flash-preview & 0.652 \\
\texttt{Forge\_\_GPT-5.4} & forgecode & gpt-5-4 & 0.818 \\
\texttt{Forge\_\_Gemini-3.1-Pro-Preview} & forge-code & gemini-3-1-pro-preview & 0.787 \\
\texttt{Forge\_\_Opus-4.6} & forgecode & claude-opus-4-6 & 0.818 \\
\texttt{Gemini\_CLI\_\_Gemini-3-Flash-Preview} & gemini-cli & gemini-3-flash-preview & 0.480 \\
\texttt{Gemini\_CLI\_\_Gemini-3.1-Pro-Preview} & gemini-cli & gemini-3-1-pro-preview & 0.608 \\
\texttt{IndusAGICodingAgent\_\_MiniMax-M2.7} & indusagi-coding-agent & minimax-m2-7 & 0.452 \\
\texttt{IndusAGICodingAgent\_\_gpt-5.3-codex} & indusagi-coding-agent & gpt-5-3-codex & 0.691 \\
\texttt{JJAgent\_\_Multiple} & jjagent & -- & 0.871 \\
\texttt{Judy\_\_Claude-Opus-4.6} & judy & claude-opus-4-6 & 0.721 \\
\texttt{Judy\_\_Gemini-3.1-Pro-Preview} & tongagents & gemini-3-1-pro-preview & 0.805 \\
\texttt{Junie\_CLI\_\_Gemini-3-Flash-Preview\ldots\textsuperscript{a}} & junie-cli & claude-opus-4-6/gemini-3-1-pr\ldots\textsuperscript{b} & 0.710 \\
\texttt{LemonCode\_\_GPT-5.3-Codex} & lemon-code & gemini-3-1-pro-preview/gpt-5-\ldots\textsuperscript{c} & 0.799 \\
\texttt{LemonHarness\_GPT-5.3-CodeX} & lemonharness & gemini-3-1-pro-preview/gpt-5-\ldots\textsuperscript{d} & 0.849 \\
\texttt{MAYA\_\_Claude-4.5-sonnet} & maya & claude-sonnet-4-5 & 0.427 \\
\texttt{MAYA\_\_Claude-4.6-opus} & maya-v2 & claude-opus-4-6 & 0.725 \\
\texttt{Meta-Harness\_\_Claude-Opus-4.6} & meta-harness & claude-opus-4-6 & 0.764 \\
\texttt{Mux\_\_Claude-Opus-4.5} & mux & claude-opus-4-5 & 0.584 \\
\texttt{Mux\_\_Claude-Opus-4.6} & mux & claude-opus-4-6 & 0.665 \\
\texttt{Mux\_\_GPT-5.2} & mux & gpt-5-2 & 0.621 \\
\texttt{Mux\_\_GPT-5.3-Codex} & mux & gpt-5-3-codex & 0.746 \\
\texttt{NexAU-AHE\_\_gpt-5.5} & nexau-ahe & gpt-5-5 & 0.847 \\
\texttt{OB-1\_GPT-5.3-Codex-Claude-Opus-4.\ldots\textsuperscript{e}} & ob-1 & claude-opus-4-5/claude-opus-4\ldots\textsuperscript{f} & 0.727 \\
\texttt{OB-1\_GPT-5.4-GPT-5.3-Codex-Claude\ldots\textsuperscript{g}} & ob-1 & claude-opus-4-5/claude-opus-4\ldots\textsuperscript{h} & 0.825 \\
\texttt{OpenCode\_\_Claude-Opus-4.5} & opencode & claude-opus-4-5 & 0.523 \\
\texttt{OpenSage\_\_GPT-5.3-Codex} & sageagent & gpt-5-3-codex & 0.794 \\
\texttt{OpenSage\_\_Gemini-3-Pro-Preview} & sageagent & gemini-3-pro-preview & 0.652 \\
\texttt{Polaris\_\_Claude-Opus-4.7-GPT-5.5-\ldots\textsuperscript{i}} & polaris & claude-opus-4-7/gemini-3-1-pr\ldots\textsuperscript{j} & 0.822 \\
\texttt{Simplai-Agent\_\_Claude-Sonnet-4.6} & simplai-agent & claude-sonnet-4-6 & 0.535 \\
\texttt{Simple-Codex\_\_GPT-5.3-Codex} & simple-codex & gpt-5-3-codex & 0.752 \\
\texttt{Terminus-KIRA\_\_Claude-Opus-4.6} & terminus-kira & claude-opus-4-6 & 0.747 \\
\texttt{Terminus-KIRA\_\_Gemini-3.1-Pro-Pre\ldots\textsuperscript{k}} & terminus-kira & gemini-3-1-pro-preview & 0.748 \\
\texttt{Terminus2\_\_Claude-Opus-4.6} & terminus2 & claude-opus-4-6 & 0.629 \\
\texttt{Terminus2\_\_DeepSeek-V3.2} & terminus2 & deepseek-v3-2 & 0.396 \\
\texttt{Terminus2\_\_GLM-4.7} & terminus2 & glm-4-7 & 0.334 \\
\texttt{Terminus2\_\_GLM-5} & terminus2 & glm-5 & 0.538 \\
\texttt{Terminus2\_\_GPT-5.3-Codex} & terminus2 & gpt-5-3-codex & 0.647 \\
\texttt{Terminus2\_\_Kimi-k2.5} & terminus2 & kimi-k2-5 & 0.433 \\
\texttt{Terminus2\_\_Minimax-m2.5} & terminus2 & minimax-m2-5 & 0.426 \\
\texttt{WOZCODE\_\_Claude-Opus-4-7} & wozcode & claude-opus-4-7 & 0.802 \\
\texttt{Wecode\_\_GPT-5.5} & wecode & gpt-5-5 & 0.881 \\
\texttt{WozCode\_\_Claude-Opus-4.6} & wozcode & claude-opus-4-6 & 0.681 \\
\texttt{cchuter\_\_minimax-m2.5} & cchuter & minimax-m2-5 & 0.427 \\
\texttt{clnkr\_\_GPT-5.5} & clnkr & gpt-5-5 & 0.661 \\
\texttt{copilot-cli\_\_claude-opus-4.6} & copilot-cli & claude-opus-4-6 & 0.607 \\
\texttt{dakou\_\_qwen3-coder-480b} & dakou-agent & qwen3-coder-480b-a35b-instruc\ldots\textsuperscript{l} & 0.272 \\
\texttt{grok-cli\_\_grok-4.20-0309-reasoning} & grok-cli & grok-4-20-0309-reasoning & 0.573 \\
\texttt{logos-latest\_\_claude-opus-4.7} & logos & -- & 0.852 \\
\texttt{logos-ts\_\_claude-opus-4.7} & logos & -- & 0.807 \\
\texttt{logos\_\_claude-opus-4.6} & logosagent & claude-opus-4-6 & 0.686 \\
\texttt{pilot-real\_\_claude-opus-4-6} & pilot & claude-opus-4-6 & 0.831 \\
\texttt{spoox-o-m\_\_GPT-5-Nano} & spoox-o-m & gpt-5-nano & 0.218 \\
\texttt{spoox-o-m\_\_GPT-5.3-Codex} & spoox-o-m & gpt-5-3-codex & 0.721 \\
\texttt{terminus-2\_\_AfterQuery-GPT-OSS-20B} & terminus2 & gpt-oss-20b-rl & 0.172 \\
\texttt{vix\_\_claude-opus-4-7} & vix & claude-opus-4-7 & 0.903 \\
\end{longtable}
\end{center}

{\scriptsize\noindent \textsuperscript{a}~\texttt{Junie\_CLI\_\_Gemini-3-Flash-Preview-Gemini-3.1-Pro-Preview-Claude-Opus-4.6-GPT-5.3-Codex} \\
\textsuperscript{b}~\texttt{claude-opus-4-6/gemini-3-1-pro-preview/gemini-3-flash-preview/gpt-5-3-codex} \\
\textsuperscript{c}~\texttt{gemini-3-1-pro-preview/gpt-5-3-codex} \\
\textsuperscript{d}~\texttt{gemini-3-1-pro-preview/gpt-5-3-codex} \\
\textsuperscript{e}~\texttt{OB-1\_GPT-5.3-Codex-Claude-Opus-4.5-Claude-Opus-4.6} \\
\textsuperscript{f}~\texttt{claude-opus-4-5/claude-opus-4-6/gpt-5-3-codex} \\
\textsuperscript{g}~\texttt{OB-1\_GPT-5.4-GPT-5.3-Codex-Claude-Opus-4.5-Claude-Opus-4.6} \\
\textsuperscript{h}~\texttt{claude-opus-4-5/claude-opus-4-6/gpt-5-3-codex/gpt-5-4} \\
\textsuperscript{i}~\texttt{Polaris\_\_Claude-Opus-4.7-GPT-5.5-Gemini-3.1-Pro} \\
\textsuperscript{j}~\texttt{claude-opus-4-7/gemini-3-1-pro-preview/gpt-5-5} \\
\textsuperscript{k}~\texttt{Terminus-KIRA\_\_Gemini-3.1-Pro-Preview} \\
\textsuperscript{l}~\texttt{qwen3-coder-480b-a35b-instruct-fp8}}

\subsection{Task panels}
The SWE-bench Verified panel consists of the following $M = 100$ instances (a fixed 20-task panel united with a seeded random draw of 80 further instances):

{\scriptsize\texttt{astropy\_\_astropy-14096, astropy\_\_astropy-7336, django\_\_django-10973, django\_\_django-11099, django\_\_django-11206, django\_\_django-11333, django\_\_django-11477, django\_\_django-11490, django\_\_django-11532, django\_\_django-11880, django\_\_django-12276, django\_\_django-12419, django\_\_django-13012, django\_\_django-13121, django\_\_django-13128, django\_\_django-13297, django\_\_django-13346, django\_\_django-13410, django\_\_django-13590, django\_\_django-13807, django\_\_django-13810, django\_\_django-13933, django\_\_django-14034, django\_\_django-14053, django\_\_django-14122, django\_\_django-14238, django\_\_django-14315, django\_\_django-14539, django\_\_django-14559, django\_\_django-14725, django\_\_django-14999, django\_\_django-15037, django\_\_django-15098, django\_\_django-15315, django\_\_django-15525, django\_\_django-15554, django\_\_django-15695, django\_\_django-15814, django\_\_django-15916, django\_\_django-15930, django\_\_django-15957, django\_\_django-16100, django\_\_django-16145, django\_\_django-16333, django\_\_django-16429, django\_\_django-16454, django\_\_django-16527, django\_\_django-16560, django\_\_django-16642, django\_\_django-16801, django\_\_django-16877, django\_\_django-9296, matplotlib\_\_matplotlib-20859, matplotlib\_\_matplotlib-23299, matplotlib\_\_matplotlib-24149, matplotlib\_\_matplotlib-24177, matplotlib\_\_matplotlib-25287, matplotlib\_\_matplotlib-25332, matplotlib\_\_matplotlib-25479, matplotlib\_\_matplotlib-25775, matplotlib\_\_matplotlib-26113, pallets\_\_flask-5014, psf\_\_requests-1724, pydata\_\_xarray-3305, pydata\_\_xarray-4075, pydata\_\_xarray-4094, pydata\_\_xarray-4966, pydata\_\_xarray-6744, pydata\_\_xarray-7229, pydata\_\_xarray-7233, pydata\_\_xarray-7393, pylint-dev\_\_pylint-6386, pylint-dev\_\_pylint-6528, scikit-learn\_\_scikit-learn-10908, scikit-learn\_\_scikit-learn-13439, sphinx-doc\_\_sphinx-10435, sphinx-doc\_\_sphinx-10466, sphinx-doc\_\_sphinx-10614, sphinx-doc\_\_sphinx-8120, sphinx-doc\_\_sphinx-8475, sphinx-doc\_\_sphinx-8593, sphinx-doc\_\_sphinx-8595, sphinx-doc\_\_sphinx-8638, sphinx-doc\_\_sphinx-9258, sphinx-doc\_\_sphinx-9367, sphinx-doc\_\_sphinx-9461, sphinx-doc\_\_sphinx-9658, sphinx-doc\_\_sphinx-9673, sphinx-doc\_\_sphinx-9698, sympy\_\_sympy-13551, sympy\_\_sympy-13647, sympy\_\_sympy-13974, sympy\_\_sympy-14248, sympy\_\_sympy-14531, sympy\_\_sympy-15875, sympy\_\_sympy-17655, sympy\_\_sympy-20154, sympy\_\_sympy-21596, sympy\_\_sympy-23950, sympy\_\_sympy-24661}}

The Terminal-Bench 2.0 panel uses all 89 tasks:

{\scriptsize\texttt{adaptive-rejection-sampler, bn-fit-modify, break-filter-js-from-html, build-cython-ext, build-pmars, build-pov-ray, caffe-cifar-10, cancel-async-tasks, chess-best-move, circuit-fibsqrt, cobol-modernization, code-from-image, compile-compcert, configure-git-webserver, constraints-scheduling, count-dataset-tokens, crack-7z-hash, custom-memory-heap-crash, db-wal-recovery, distribution-search, dna-assembly, dna-insert, extract-elf, extract-moves-from-video, feal-differential-cryptanalysis, feal-linear-cryptanalysis, filter-js-from-html, financial-document-processor, fix-code-vulnerability, fix-git, fix-ocaml-gc, gcode-to-text, git-leak-recovery, git-multibranch, gpt2-codegolf, headless-terminal, hf-model-inference, install-windows-3.11, kv-store-grpc, large-scale-text-editing, largest-eigenval, llm-inference-batching-scheduler, log-summary-date-ranges, mailman, make-doom-for-mips, make-mips-interpreter, mcmc-sampling-stan, merge-diff-arc-agi-task, model-extraction-relu-logits, modernize-scientific-stack, mteb-leaderboard, mteb-retrieve, multi-source-data-merger, nginx-request-logging, openssl-selfsigned-cert, overfull-hbox, password-recovery, path-tracing, path-tracing-reverse, polyglot-c-py, polyglot-rust-c, portfolio-optimization, protein-assembly, prove-plus-comm, pypi-server, pytorch-model-cli, pytorch-model-recovery, qemu-alpine-ssh, qemu-startup, query-optimize, raman-fitting, regex-chess, regex-log, reshard-c4-data, rstan-to-pystan, sam-cell-seg, sanitize-git-repo, schemelike-metacircular-eval, sparql-university, sqlite-db-truncate, sqlite-with-gcov, torch-pipeline-parallelism, torch-tensor-parallelism, train-fasttext, tune-mjcf, video-processing, vulnerable-secret, winning-avg-corewars, write-compressor}}

\end{document}